\documentclass[10pt,twocolumn,letterpaper]{article}

\usepackage[pagenumbers]{cvpr} %

\usepackage{graphicx}
\usepackage{amsmath}
\usepackage{amssymb}
\usepackage{booktabs}

\usepackage{algorithm}
\usepackage{algpseudocode}
\usepackage{amsmath}
\usepackage{amssymb}
\usepackage{booktabs}
\usepackage{dsfont}
\usepackage{graphicx}
\usepackage{makecell}
\usepackage{multirow}
\usepackage{pifont}%
\usepackage[group-separator={,}]{siunitx}
\usepackage{tabularx}
\usepackage{xcolor}
\usepackage{caption}
\usepackage{soul}
\usepackage{changepage,threeparttable}
\usepackage{colortbl}
\usepackage[accsupp]{axessibility}  %
\usepackage{mathtools}
\usepackage{xfrac}
\usepackage{catchfile}
\usepackage{longtable}
\usepackage{tabu}
\usepackage{supertabular}
\usepackage{multicol}

\newcolumntype{Y}{>{\centering\arraybackslash}X}

\renewcommand{\ie}{i.e.,\,}

\DeclareMathOperator*{\argmin}{arg\,min}

\definecolor{color_1}{RGB}{255,0,128}
\definecolor{color_2}{RGB}{128,128,0}
\definecolor{color_3}{RGB}{0,128,0}
\definecolor{color_4}{RGB}{128,0,0}
\definecolor{color_5}{RGB}{128,0,128}
\definecolor{cadetgrey}{RGB}{0.57, 0.64, 0.69}
\definecolor{color_red}{RGB}{255,0,0}
\definecolor{color_green}{RGB}{0,255,0}
\definecolor{color_blue}{RGB}{0,0,255}
\definecolor{color_gray}{RGB}{127,127,127}
\newcommand\Red[1] {\emph{\textcolor{color_red}{#1}}}
\newcommand\Green[1] {\emph{\textcolor{color_green}{#1}}}
\newcommand\Blue[1] {\emph{\textcolor{color_blue}{#1}}}
\newcommand\Gray[1] {\emph{\textcolor{color_gray}{#1}}}

\newcommand{\projname}[0]{\textsc{APAP}}

\newcommand{\eqn}[0]{Eqn.}
\newcommand{\fig}[0]{Fig.}
\newcommand{\tab}[0]{Tab.}
\newcommand{\alg}[0]{Alg.}
\newcommand{\shortsec}[0]{Sec.}

\newcommand*{\APAP}[0]{\textbf{APAP}}
\newcommand*{\suppl}[0]{the \textbf{supplementary material}}
\newcommand{\benchmark}[0]{\textsc{APAP-Bench}}
\newcommand{\benchmarkTwoD}[0]{\textsc{APAP-Bench 2D}}
\newcommand{\benchmarkThreeD}[0]{\textsc{APAP-Bench 3D}}
\newcommand{\FirstStage}[0]{\texttt{FirstStage}}
\newcommand{\SecondStage}[0]{\texttt{SecondStage}}

\definecolor{cvprblue}{rgb}{0.21,0.49,0.74}
\usepackage[pagebackref,breaklinks,colorlinks,citecolor=cvprblue]{hyperref}

\usepackage[capitalize]{cleveref}
\crefname{section}{Sec.}{Secs.}
\Crefname{section}{Section}{Sections}
\Crefname{table}{Table}{Tables}
\crefname{table}{Tab.}{Tabs.}

\def\paperID{3020} %
\def\confName{CVPR}
\def\confYear{2024}

\begin{document}

\title{As-Plausible-As-Possible: Plausibility-Aware Mesh Deformation
\\Using 2D Diffusion Priors}

\author{
Seungwoo Yoo\textsuperscript{*1} \quad
Kunho Kim\textsuperscript{*1} \quad
Vladimir G. Kim\textsuperscript{2} \quad
Minhyuk Sung\textsuperscript{1} \\[0.2em]
\textsuperscript{1}KAIST $\quad$ \textsuperscript{2}Adobe Research
}

\twocolumn[{%
\renewcommand\twocolumn[1][]{#1}%

\maketitle
\vspace{-40.0pt}
\begin{center}
    \captionsetup{type=figure, labelfont=bf}
    \includegraphics[width=\textwidth]{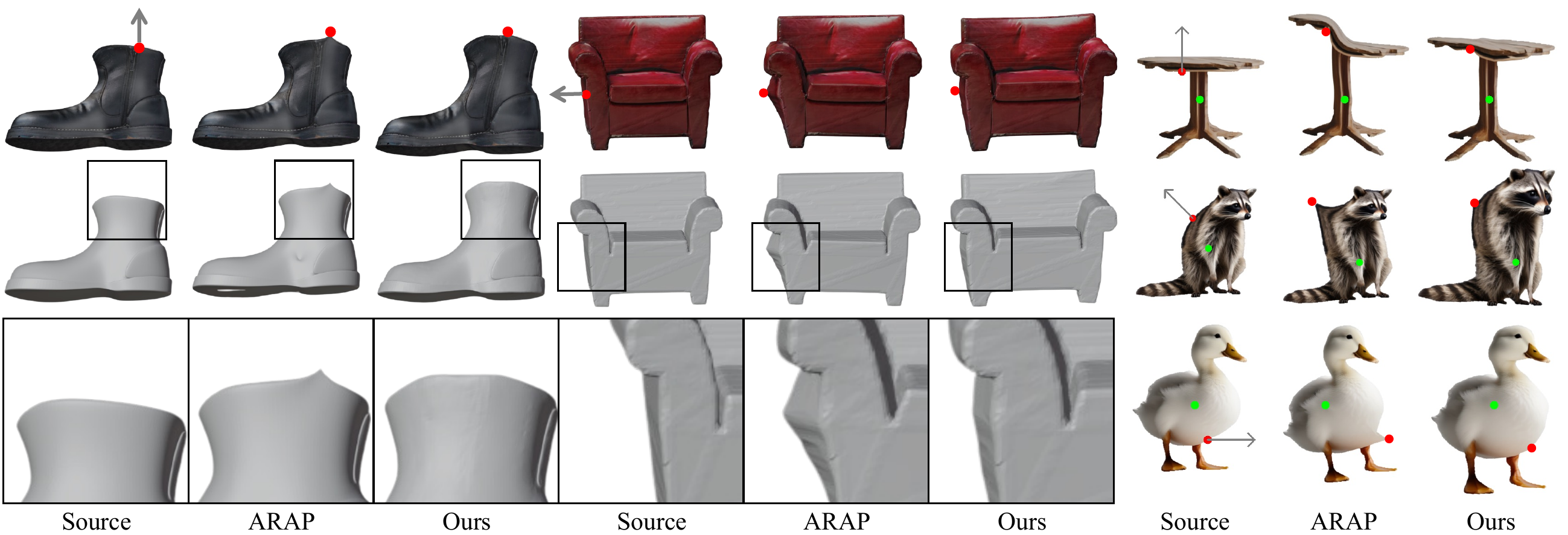}
    \vspace{-20pt}
    \caption{
    \APAP{}, our novel shape deformation method, enables plausibility-aware mesh deformation and preservation of fine details of the original mesh
    offering an interface that alters geometry by directly displacing a handle (\Red{red}) along a direction (\Gray{gray}), fixing an anchor vertex (\Green{green}).
    Using a diffusion prior results in smoother geometry around the armchair handle, as seen in the example (middle column).
    }
    \label{fig:teaser}
\end{center}
}]

\def\thefootnote{*}\footnotetext{Equal contribution.}\def\thefootnote{\arabic{footnote}}

\begin{abstract}
\vspace{-10pt}
We present As-Plausible-as-Possible (APAP) mesh deformation technique that leverages 2D diffusion priors to preserve the plausibility of a mesh under user-controlled deformation. 
Our framework uses per-face Jacobians to represent mesh deformations, where mesh vertex coordinates are computed via a differentiable Poisson Solve. The deformed mesh is rendered, and the resulting 2D image is used in the Score Distillation Sampling (SDS) process, which enables extracting meaningful plausibility priors from a pretrained 2D diffusion model. To better preserve the identity of the edited mesh, we fine-tune our 2D diffusion model with LoRA. Gradients extracted by SDS and a user-prescribed handle displacement are then backpropagated to the per-face Jacobians, and we use iterative gradient descent to compute the final deformation that balances between the user edit and the output plausibility. 
We evaluate our method with 2D and 3D meshes and demonstrate qualitative and quantitative improvements when using plausibility priors over geometry-preservation or distortion-minimization priors used by previous techniques. 
Our project page is at: \href{https://as-plausible-as-possible.github.io/}{https://as-plausible-as-possible.github.io/}

\end{abstract}

\section{Introduction}
\label{sec:intro}

For 2D and 3D content, mesh is the most prevalent representation, thanks to its efficiency in storage, simplicity in rendering and also compatibility in common graphics pipelines, versatility in diverse applications such as design, physical simulation, and 3D printing, and flexibility in terms of decomposing geometry and appearance information, with widespread adoption in the industry.

For the creation of 2D and 3D meshes, recent breakthroughs in generative models~\cite{Shen:2021DMTet, Shue:2023TriplaneDiffusion, Koo:2023SALAD, Poole:2023Dreamfusion, Wang2023:ProlificDreamer,Shi2023:MVDream, Liu2023:SyncDreamer,Tang2023:DreamGaussian}
have demonstrated significant advances. These breakthroughs enable users to easily generate content from a text prompt~\cite{Poole:2023Dreamfusion, Wang2023:ProlificDreamer,Shi2023:MVDream, Liu2023:SyncDreamer,Tang2023:DreamGaussian}, or from photos~\cite{Shi2023:MVDream, Raj2023:Dreambooth3D}. 
However, visual content creation typically involves numerous editing processes, deforming the content to satisfy users' desires through interactions such as mouse clicks and drags.
Facilitating such interactive editing has remained relatively underexplored in the context of recent generative techniques.

Mesh deformation is a subject that has been researched for decades in computer graphics. Over time, researchers have established well-defined methodologies, characterizing mesh deformation as an optimization problem that aims to preserve specific geometric properties, such as the Mesh Laplacian~\cite{Lipman:2004Differential,Sorkine:2004LaplacianSurfaceEditing,Lipman:2005RotationInvariant}, local rigidity~\cite{Igarashi:2005ARAP,Sorkine:2007ARAP}, and mesh surface Jacobians~\cite{Aigerman:2022NJF, Gao:2023TextDeformer}, while satisfying given constraints. To facilitate user interaction, these methodologies have been extended to introduce specific user-interactive deformation handles, such as keypoints~\cite{Jacobson:2011BoundedBiharmonics, Wang:2015Biharmonic, Kim:2023OptCtrlPoints}, cage mesh~\cite{Ju:2005MVC,Joshi:2007HarmonicCoordinates,Lipman:2008GreenCoordinates,Weber:2009Barycentric, Yifan:2020NeuralCage, Jakab:2021KeypointDeformer}, and skeleton~\cite{Baran:2007AutomaticRigging, Zhan:2020RigNet, Xu:2022Morig}, with the blending functions defined based on the preservation of geometric properties.

Despite the widespread use of classical mesh deformation methods, they often fail to meet users' needs because they do not incorporate the perceptual plausibility of the outputs. For example, as illustrated in \fig~\ref{fig:teaser}, when a user intends to drag a point on the top of a table image, the classical deformation technique may introduce unnatural bending instead of lifting the tabletop. This limitation arises because deformation techniques solely based on geometric properties do not incorporate such semantic and perceptual priors, resulting in the mesh editing process becoming more tedious and time-consuming.

Recent learning-based mesh deformation techniques~\cite{Yifan:2020NeuralCage,Jakab:2021KeypointDeformer,Kim:2023OptCtrlPoints,Zhan:2020RigNet,Liu:2021DeepMetaHandles,Aigerman:2022NJF,Tang:2022NSDP} have attempted to address this problem in a data-driven way. However, they are also limited by relying on the existence of certain variations in the training data. Even recent large-scale 3D datasets~\cite{Chang:2015Shapenet, Wu:2023OmniObject3D, Deitke:2023Objaverse, Deitke:2023ObjaverseXL} have not reached the scale that covers all possible visual content users might intend to create.

To this end, we introduce our novel mesh deformation framework, dubbed APAP (As-Plausible-As-Possible), which exploits 2D image priors from a diffusion model pretrained on an Internet-scale image dataset to enhance the plausibility of deformed 2D and 3D meshes while preserving the geometric priors of the given shape. Recently, score distillation sampling (SDS)~\cite{Poole:2023Dreamfusion} has demonstrated great success in generating plausible 2D and 3D content, such as NeRF~\cite{Zhuang:2023DreamEditor, Kim:2023CSD, Park:2023EDNeRF} and vector images~\cite{Jain:2023VectorFusion, Iluz:2023Typography}, using the distilled 2D image priors from a diffusion model. We incorporate these diffusion-model-based 2D priors into the optimization-based deformation framework, achieving the best synergy between geometry-based optimization and distilled-prior-based optimization.

To achieve this optimal synergy between geometric and perceptual priors within a unified framework, we introduce an alternative optimization approach. At each step, we first update the Jacobian of each mesh face using the SDS loss and user-provided constraints. Subsequently, the mesh vertex positions are recalculated by solving Poisson's equation with the updated face Jacobians. The direct application of the 2D diffusion prior via SDS, however, tends to compromise the identity of the given objects—an essential aspect in deformation.
We thus enhance the identity awareness of the diffusion prior by finetuning it with the provided source image. The model is integrated into our two-stage pipeline that initiates deformation without the perceptual prior (SDS) and refines it with SDS and the given constraints afterward to create deformations that adhere to user-defined editing instructions while remaining visually plausible.

In experiments, we examine \APAP{} using \benchmark{} consisting of 3D and 2D triangular meshes and editing instructions. The proposed method produces plausible deformations of 3D meshes compared to its baseline~\cite{Sorkine:2007ARAP} based exclusively on a geometric prior. Evaluation in the task of 2D mesh editing further verifies the effectiveness of \APAP{} as illustrated by the highest $k$-NN GIQA score~\cite{Gu:2020GIQA} in quantitative analysis, and the higher preference over the baseline in a user study.

\section{Related Work}
\label{sec:related_work}

\subsection{Geometric Mesh Deformation}

Mesh deformation has been one of the central problems in geometry processing and is thus addressed by a wide range of techniques.
Cage-based methods~\cite{Ju:2005MVC,Joshi:2007HarmonicCoordinates,Lipman:2008GreenCoordinates,Weber:2009Barycentric} let users alter meshes by manipulating cages enclosing them, calculating a point inside as a weighted sum of cage vertices.
Skeleton-based approaches~\cite{Weber:2007ContextAware,Baran:2007AutomaticRigging,Zhan:2020RigNet, Xu:2022Morig} offer animation control by mapping surface points to underlying joints and bones, ideal for animating human/animal-like figures.
Unlike the previous techniques that require the manual cage or skeleton construction, biharmonic coordinates-based methods~\cite{Jacobson:2011BoundedBiharmonics,Wang:2015Biharmonic} automate establishing mappings from control points to vertices by formulating optimization problems.
Other types of works instead allow users to manipulate shapes via direct vertex displacement while imposing constraints on local surface geometry, including rigidity~\cite{Igarashi:2005ARAP,Sorkine:2007ARAP} and Laplacian smoothness~\cite{Lipman:2004Differential,Sorkine:2004LaplacianSurfaceEditing,Lipman:2005RotationInvariant}.
Such hand-crafted deformation priors often lack consideration of visual plausibility, necessitating careful control point placement and iterative manual refinement to achieve satisfactory results.

\subsection{Data-Driven Mesh Deformation}

Data-driven approaches to mesh  deformation~\cite{Yifan:2020NeuralCage,Jakab:2021KeypointDeformer,Kim:2023OptCtrlPoints,Zhan:2020RigNet,Liu:2021DeepMetaHandles,Aigerman:2022NJF,Tang:2022NSDP} learn from shape collections, utilizing neural networks to infer parameters for classical deformation techniques, such as cage vertex coordinates and displacements~\cite{Yifan:2020NeuralCage}, keypoints~\cite{Jakab:2021KeypointDeformer,Wang:2015Biharmonic,Kim:2023OptCtrlPoints}, subspaces of keypoint arrangements~\cite{Liu:2021DeepMetaHandles}, differential coordinates~\cite{Aigerman:2022NJF}, etc. However, these methods assume the availability of large-scale category-specific shape collection~\cite{Yifan:2020NeuralCage,Jakab:2021KeypointDeformer,Wang:2015Biharmonic,Kim:2023OptCtrlPoints,Zhan:2020RigNet} or require dense correspondences between them~\cite{Tang:2022NSDP,Aigerman:2022NJF}, limiting their applicability to new, out-of-sample shapes. We instead propose to directly mine deformation priors from pretrained diffusion models. Leveraging a generic (category-agnostic) image generative model trained on an Internet-scale image dataset, we devise a method that easily generalizes to novel 2D and 3D shapes while lifting the requirement for shape collections.

\subsection{Pretrained 2D Priors for Shape Manipulation}

Image analysis~\cite{Radford:2021CLIP} and generation~\cite{Rombach:2022LDM,Zhang2023ControlNet, Bar2023MultiDiffusion, lee2023SyncDiffusion} techniques can serve as effective visual priors for image editing tasks~\cite{Cao:2023MasaCtrl, Hertz:2023P2P, Tumanyan:2023PnP, Zhang:2023RIVAL,Shi:2023DragDiffusion}. In addition, recent work~\cite{Gal:2022Textual,Ruiz:2022DreamBooth} and their adaption~\cite{HF:DreamBoothLoRA}, enable personalized image generation and editing by learning a text embedding~\cite{Gal:2022Textual} or fine-tuning additional parameters, such as LoRA~\cite{Hu:2022LoRA} to preserve and replicate the identities of given exemplars during editing. One interesting work is DragDiffusion~\cite{Shi:2023DragDiffusion}, akin to DragGAN~\cite{Pan:2023DragGAN}, which introduces a drag-based user interface for image editing through the manipulation of latent representations. However, it is not extendable to the deformation of parametric images, such as 2D meshes, and also 3D shapes. Another interesting line of works ~\cite{Michel:2022Text2Mesh,Khalid:2022CLIPMesh,Gao:2023TextDeformer} extends the idea further to manipulate shapes by propagating image-based gradients to the underlying shape representations. They maximize CLIP~\cite{Radford:2021CLIP} similarity between the renderings and text prompts to either add geometric textures~\cite{Michel:2022Text2Mesh}, jointly update both vertices and texture~\cite{Khalid:2022CLIPMesh}, or deform a shape parameterized by per-triangle Jacobians~\cite{Gao:2023TextDeformer}. 
In contrast to such text-driven editing techniques, we build on Score Distillation Sampling (SDS)~\cite{Poole:2023Dreamfusion} to enable direct manipulation of shapes via handle displacement, ensuring visual plausibility. While the technique is prevalent in various problems ranging from text-to-3D~\cite{Poole:2023Dreamfusion,Wang2023:ProlificDreamer,Shi2023:MVDream,Liu2023:SyncDreamer,Tang2023:DreamGaussian}, image editing~\cite{Hertz:2023DDS} and neural field editing~\cite{Zhuang:2023DreamEditor}, it has not been adopted for shape deformation.

\section{Method}
\label{sec:method}

We present \APAP{}, a novel handle-based mesh deformation framework capable of producing visually plausible deformations of either 2D or 3D triangular meshes.
To achieve this goal, we integrate powerful 2D diffusion priors into a learnable Jacobian field representation of shapes.

We emphasize that leveraging 2D priors, such as latent diffusion models (LDMs)~\cite{Rombach:2022LDM} trained on large-scale datasets~\cite{Schuhmann:2022Laion5B}, for shape deformation poses challenges that require meticulous design choices.
The following sections will delve into the details of shape representation (\shortsec{}~\ref{subsec:deform_param}) and diffusion prior (\shortsec{}~\ref{subsec:score\string_distillation}), offering a rationale for the design decisions underpinning our framework (\shortsec{}~\ref{subsec:as_plausible_as_possible}).

\subsection{Representing Shapes as Jacobian Fields}
\label{subsec:deform_param}

Let $\mathcal{M}_0 = \left(\mathbf{V}_0, \mathbf{F}_0\right)$ denote a source mesh to be deformed, represented by vertices $\mathbf{V}_0 \in \mathbb{R}^{V \times 3}$ and faces $\mathbf{F}_0 \in \mathbb{R}^{F \times 3}$. 
Users are allowed to select a set of vertices used as movable handles designated by an indicator matrix $\mathbf{K}_h \in \{0, 1\}^{V_h \times V}$. We also require users to select a set of anchors, represented as another indicator matrix $\mathbf{K}_a \in \{0, 1\}^{V_a \times V}$, to avoid trivial solutions (\ie global translations).
Then, the handle and anchor vertices become $\mathbf{V}_h = \mathbf{K}_h \mathbf{V}_0$ and $\mathbf{V}_a = \mathbf{K}_a \mathbf{V}_0$.

Our framework also expects a set of vectors $\mathbf{D}_h \in \mathbb{R}^{V_h \times 3}$ that indicate the directions along which the handles will be displaced. 
Furthermore, we let $\mathbf{T}_h = \mathbf{V}_h + \mathbf{D}_h$ and $\mathbf{T}_a = \mathbf{V}_a$ denote the target positions of the user-specified handles and anchors, respectively.

In this work, we employ a Jacobian field $\mathbf{J}_0 = \{\mathbf{J}_{0, f} \vert f \in \mathbf{F}_0 \}$, a dual representation of $\mathcal{M}_0$, defined as a set of per-face Jacobians $\mathbf{J}_{0,f} \in \mathbb{R}^{3 \times 3}$ where
\begin{align}
    \mathbf{J}_{0,f} &= \boldsymbol{\nabla}_f \mathbf{V}_0,
\end{align}
and $\boldsymbol{\nabla}_f$ is the gradient operator of triangle $f$.

Conversely, we compute a set of \textit{deformed} vertices $\mathbf{V}^{*}$ from a given Jacobian field $\mathbf{J}$ by solving a Poisson's equation
\begin{align}
    \mathbf{V}^{*} = \argmin_{\mathbf{V}} \Vert \mathbf{L} \mathbf{V} - \boldsymbol{\nabla}^{T} \mathcal{A} \mathbf{J} \Vert^{2},
    \label{eqn:poisson_unconstrained}
\end{align}
where $\boldsymbol{\nabla}$ is a stack of per-face gradient operators, $\mathcal{A} \in \mathbb{R}^{3F \times 3F}$ is the mass matrix and $\mathbf{L} \in \mathbb{R}^{V \times V}$ is the cotangent Laplacian of $\mathcal{M}_0$, respectively. Since $\mathbf{L}$ is rank-deficient, the solution of \eqn{}~\ref{eqn:poisson_unconstrained} cannot be uniquely determined unless we impose constraints. We thus consider a constrained optimization problem
\begin{align}
    \mathbf{V}^{*} = \argmin_{\mathbf{V}} \Vert \mathbf{L} \mathbf{V} - \boldsymbol{\nabla}^{T} \mathcal{A} \mathbf{J} \Vert^{2} + \lambda \Vert \mathbf{K}_a \mathbf{V} - \mathbf{T}_a \Vert^{2},
    \label{eqn:poisson_constrained}
\end{align}
where $\lambda \in \mathbb{R}^{+}$ is a weight for the constraint term. Note that we solve \eqn{}~\ref{eqn:poisson_constrained} with the user-specified anchors as constraints to determine $\mathbf{V}^*$.

Taking the derivative with respect to $\mathbf{V}$, the problem in \eqn{}~\ref{eqn:poisson_constrained} turns into a system of equations
\begin{align}
    \left(\mathbf{L}^{T} \mathbf{L} + \lambda \mathbf{K}_a^{T}\mathbf{K}_a \right) \mathbf{V} = \mathbf{L}^{T} \boldsymbol{\nabla}^{T} \mathcal{A} \mathbf{J} + \lambda \mathbf{K}_a^{T} \textbf{T}_a,
    \label{eqn:poisson_constrained_linear}
\end{align}
which can be efficiently solved using a differentiable solver~\cite{Aigerman:2022NJF} implementing Cholesky decomposition.

We let $g$ denote a functional representing the aforementioned differentiable solver for notational convenience, $\mathbf{V}^* = g\left(\mathbf{J}, \mathbf{K}_a, \mathbf{T}_a \right)$.
Since $g$ is differentiable, we can deform $\mathcal{M}_0$ by propagating upstream gradients from various loss functions to the underlying parameterization $\mathbf{J}$. For instance, one may impose a \textit{soft} constraint on the locations of selected handles during optimization with the objective of the form:
\begin{align}
\mathcal{L}_h = \Vert \mathbf{K}_h \mathbf{V}^* - \mathbf{T}_h \Vert^2.
\label{eqn:loss_h}
\end{align}
We will discuss how such a soft constraint can be blended into our framework in \shortsec{}~\ref{subsec:as_plausible_as_possible}.
Next, we describe how to incorporate a pretrained diffusion model as a prior for visual plausibility.

\begin{figure*}[!t]
\captionsetup{type=figure, labelfont=bf}
\centering
\includegraphics[width=\textwidth]{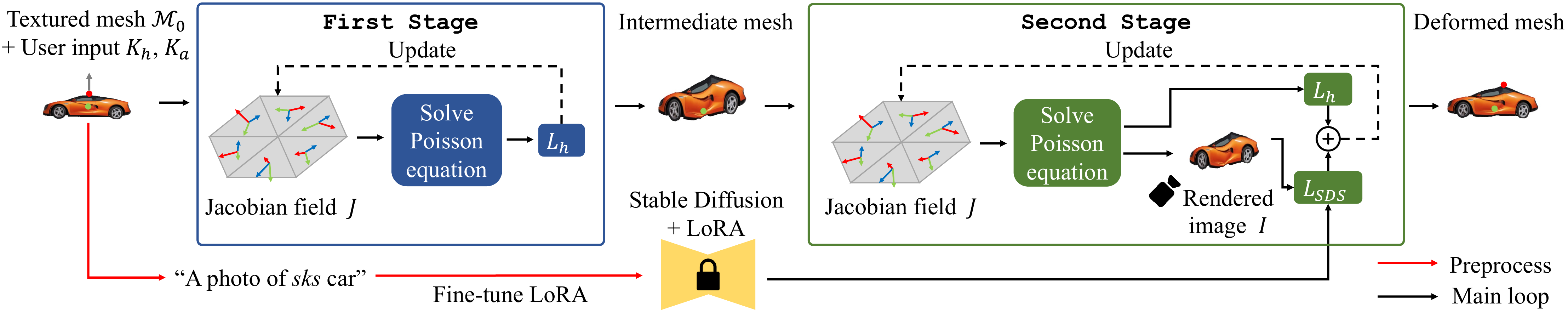}
\vspace{-20pt}
\caption{
\textbf{The overview of \projname{}.} \projname{} parameterizes a triangular mesh as a per-face Jacobian field that can be updated via gradient-descent. Given a textured mesh and user inputs specifying the handle(s) and anchor(s), our framework initializes a Jacobian field as a trainable parameter. During the first stage, the Jacobian field is updated via iterative optimization of $\mathcal{L}_h$, a soft constraint that initially deforms the shape according to the user's instruction. In the following stage, the mesh is rendered using a differentiable renderer $\mathcal{R}$ and the rendered image is provided as an input to a diffusion prior finetuned with LoRA~\cite{Hu:2022LoRA} that computes the SDS loss $\mathcal{L}_{\text{SDS}}$. The joint optimization of $\mathcal{L}_{h}$ and $\mathcal{L}_{\text{SDS}}$ improves the visual plausibility of the mesh while conforming to the given edit instruction.
}
\label{fig:method}
\end{figure*}

\subsection{Score Distillation for Shape Deformation}
\label{subsec:score_distillation}

While traditional mesh deformation techniques make variations that match the given \textit{geometric} constraints, their lack of consideration on \textit{visual plausibility} results in unrealistic shapes. %
Motivated by recent success in text-to-3D literature, we harness a powerful 2D diffusion prior~\cite{Rombach:2022LDM} in our framework as a critic that directs deformation by scoring the realism of the current shape.

Specifically, we distill its prior knowledge via Score Distillation Sampling (SDS)~\cite{Poole:2023Dreamfusion}. Let $\mathbf{J}$ denote the current Jacobian field and $\mathbf{V}^*$ be the set of vertices computed from $\mathbf{J}$ following the procedure described in \shortsec{}~\ref{subsec:deform_param}.

We render $\mathcal{M}^* = \left(\mathbf{V}^*, \mathbf{F} \right)$ from a viewpoint defined by camera extrinsic parameters $\mathbf{C}$ using a differentiable renderer $\mathcal{R}$, producing an image $\mathcal{I} = \mathcal{R}\left(\mathcal{M}^*, \mathbf{C}\right)$. The diffusion prior $\hat{\epsilon}_{\phi}$ then rates the realism of $\mathcal{I}$, producing a gradient
\begin{align}
    \nabla_{\mathbf{J}} \mathcal{L}_{\text{SDS}} \left(\phi, \mathcal{I}\right) = \mathbb{E}_{t, \epsilon} \left[w\left(t\right) \left(\hat{\epsilon}_{\phi}\left(\mathbf{z}_t; y, t\right) - \epsilon \right) \frac{\partial \mathcal{I}}{\partial \mathbf{J}}\right],
    \label{eqn:sds_loss}
\end{align}
where $t \sim \mathcal{U}\left(0, 1\right)$, $\epsilon \sim \mathcal{N}\left(\mathbf{0}, \mathbf{I}\right)$, and $\mathbf{z}_t$ is a noisy latent embedding of $\mathcal{I}$. The propagated gradient alters the geometry of $\mathcal{M}$ by modifying $\mathbf{J}$.

To increase the instance-awareness of the diffusion model, we follow recent work~\cite{Ruiz:2022DreamBooth,Shi:2023DragDiffusion} on personalized image editing and finetune the model using LoRA~\cite{Hu:2022LoRA}. %
In particular, we first render $\mathcal{M}$ from $n$ different viewpoints to obtain a set $\mathcal{I} = \{\mathcal{I}_1, \dots, \mathcal{I}_n\}$ of training images and inject additional parameters to the model, resulting in an expanded set of network parameters $\phi^{\prime}$.
The parameters are then optimized with a denoising loss~\cite{Rombach:2022LDM}
\begin{align}
    \mathcal{L} = \mathbb{E}_{t, \epsilon, \mathbf{z}} \left[ \Vert \hat{\epsilon}_{\phi^{\prime}} \left(\mathbf{z}_t; y, t\right) - \epsilon \Vert^2 \right],
\end{align}
where $\mathbf{z}_t$ denotes a latent of a training image perturbed with noise at timestep $t$.

The finetuned diffusion prior, together with a learnable Jacobian field representation of the source mesh $\mathcal{M}_0$, comprises the proposed framework described in the following section.

\subsection{As-Plausible-As-Possible (APAP)}
\label{subsec:as_plausible_as_possible}

\begin{algorithm}
\caption{As-Plausible-As-Possible}
\label{alg:apap_main_loop}
\begin{algorithmic}
\State \textbf{Parameters:} $g$, $\mathcal{R}$, $\phi$, $\gamma$, $M$, $N$
\State \textbf{Inputs:} $\mathcal{M}_0 = \left(\mathbf{V}_0, \mathbf{F}_0\right)$, $\mathbf{K}_{a}$, $\mathbf{K}_{h}$, $\mathbf{T}_{a}$, $\mathbf{T}_{h}$, $\left\{ \mathbf{C}_i \right\}_{i=1}^n$
\State \textbf{Output:} $\mathcal{M}$

\\
\Procedure{FirstStage}{$\mathbf{J}$, $\mathbf{K}_a$, $\mathbf{K}_{h}$, $\mathbf{T}_a$, $\mathbf{T}_h$, $g$}
    \For{$i=1,2,\dots,M$}
        \State{$\mathbf{V}^*$ $\gets$ $g\left(\mathbf{J}, \mathbf{K}_a, \mathbf{T}_a\right)$} \Comment{Solving \eqn{}~\ref{eqn:poisson_constrained_linear}}
        \State{$\mathbf{J}$ $\gets$ $\mathbf{J} - \gamma \nabla_{\mathbf{J}}\mathcal{L}_h \left(\mathbf{V}^{*}, \mathbf{K}_h, \mathbf{T}_h \right)$}
    \EndFor
    \State{\Return $\mathbf{J}$}
\EndProcedure

\Procedure{SecondStage}{$\mathbf{J}$, $\mathbf{F}_0$, $\mathbf{K}_a$, $\mathbf{K}_{h}$, $\mathbf{T}_a$, $\mathbf{T}_h$, $g$, $\phi$, $\left\{ \mathbf{C}_i \right\}$}
    \For{$i=1,2,\dots,N$}
        \State{$\mathbf{V}^*$ $\gets$ $g\left(\mathbf{J}, \mathbf{K}_a, \mathbf{T}_a\right)$} \Comment{Solving \eqn{}~\ref{eqn:poisson_constrained_linear}}
        \State{$\mathcal{M}^*$ $\gets$ $\left(\mathbf{V}^*, \mathbf{F}_0\right)$}
        \State{$\mathbf{C}$ $\sim$ $\mathcal{U}$($\left\{ \mathbf{C}_i \right\}$)} \Comment{Viewpoint Sampling}
        \State{$\mathcal{I}$ $\gets$ $\mathcal{R}\left(\mathcal{M}^*, \mathbf{C}\right)$} \Comment{Rendering}
        \State{$\mathbf{J}$ $\gets$ $\mathbf{J} - \gamma \nabla_{\mathbf{J}} \left(\mathcal{L}_{\text{SDS}} \left(\phi, \mathcal{I} \right) + \mathcal{L}_h \left(\mathbf{V}^{*}, \mathbf{K}_h, \mathbf{T}_h\right) \right)$}
    \EndFor
    \State{\Return $\mathbf{J}$}
\EndProcedure

\\
\State{$\phi$ $\gets$ \Call{LoRA}{$\phi$, $\mathcal{M}_0$, $\mathcal{R}$, $\left\{\mathbf{C}_i\right\}$}}
\State{$\mathbf{J}$ $\gets$ $\{\mathbf{J}_{0, f} \vert f \in \mathbf{F}_0 \}$}
\State{$\mathbf{J}$ $\gets$ \Call{FirstStage}{$\mathbf{J}$, $\mathbf{K}_a$, $\mathbf{K}_{h}$, $\mathbf{T}_a$, $\mathbf{T}_h$, $g$}}
\State{$\mathbf{J}$ $\gets$ \Call{SecondStage}{$\mathbf{J}$, $\mathbf{F}_0$, $\mathbf{K}_a$, $\mathbf{K}_{h}$, $\mathbf{T}_a$, $\mathbf{T}_h$, $g$, $\phi$, $\left\{ \mathbf{C}_i \right\}$}}
\State{$\mathbf{V}$ $\gets$ $g\left(\mathbf{J}, \mathbf{K}_a, \mathbf{T}_a \right)$}
\State{$\mathcal{M}$ $\gets$ $\left(\mathbf{V}, \mathbf{F}_0\right)$}
\State{\Return $\mathcal{M}$}
\end{algorithmic}
\end{algorithm}

\APAP{} tackles the problem of plausibility-aware shape deformation by harmonizing the best of both worlds: a learnable shape representation founded on classical geometry processing, robust to noisy gradients, and a powerful 2D diffusion prior finetuned with the image(s) of the source mesh for better instance-awareness.

We provide an overview of the proposed pipeline in \fig~\ref{fig:method} and the algorithm in \alg~\ref{alg:apap_main_loop}. We will delve into details in the following. %
Provided with a textured mesh $\mathcal{M}_0$, handles $\mathbf{K}_h$, anchors $\mathbf{K}_a$, as well as their target positions $\mathbf{T}_h$ and $\mathbf{T}_a$ as inputs, \APAP{} yields a plausible deformation $\mathcal{M}$ of $\mathcal{M}_0$ that conforms to the given handle-target constraints. Before deforming $\mathcal{M}_0$, we render $\mathcal{M}_0$ from a single view in the case of 2D meshes and four canonical views (\ie front, back, left, and right) for 3D meshes and use the images to finetune Stable Diffusion~\cite{Rombach:2022LDM} by optimizing LoRA~\cite{Hu:2022LoRA} parameters injected to the model (the \Red{red} line in \fig~\ref{fig:method}). Simultaneously, \APAP{} computes the Jacobian field $\mathbf{J}_0$ of the input mesh $\mathcal{M}_0$ and initializes it as a trainable parameter $\mathbf{J}$.

\APAP{} deforms the input mesh through two stages. In the \FirstStage{}, it first deforms the input mesh according to instructions from users without taking visual plausibility into account. The subsequent \SecondStage{} integrates a 2D diffusion prior into the optimization loop, simultaneously enforcing user constraints and visual plausibility.

At every iteration of the \FirstStage{} illustrated as the \Blue{blue} box in \fig~\ref{fig:method}, we compute the vertex positions $\mathbf{V}^*$ corresponding to the current Jacobian field $\mathbf{J}$ by solving \eqn~\ref{eqn:poisson_constrained} using the anchors specified by $\mathbf{K}_a$ as hard constraints. Then, we compute the soft constraint $\mathcal{L}_h$ defined as \eqn~\ref{eqn:loss_h} that drags a set of handle vertices $\mathbf{K}_h \mathbf{V}^{*}$ toward the corresponding targets $\mathbf{T}_h$. The interleaving of differentiable Poisson solve and optimization of $\mathcal{L}_h$ via gradient-descent is repeated for $M$ iterations. This progressively updates $\mathbf{J}$, treated as a learnable black box in our framework, deforming $\mathcal{M}_0$.
Consequently, the edited mesh $\mathcal{M}^*=\left(\mathbf{J}, \mathbf{F}_0\right)$ follows user constraints at the cost of the degraded plausibility, mitigated in the following stage through the incorporation of a diffusion prior.

The result of \FirstStage{} then serves as an initialization for the \SecondStage{}, illustrated as the \Green{green} box in \fig~\ref{fig:method} guided by plausibility constraint $\mathcal{L}_{\text{SDS}}$. Unlike the \FirstStage{} where the update of $\mathbf{J}$ was purely driven by the geometric constraint $\mathcal{L}_h$, we aim to steer the optimization based on the visual plausibility of the current mesh $\mathcal{M}^*$. To achieve this, we render $\mathcal{M}^*$ using a differentiable renderer $\mathcal{R}$ using the same viewpoint(s) from which the training image(s) for finetuning was rendered. When deforming 3D meshes, we randomly sample one viewpoint at each iteration. The rendered image $\mathcal{I}$ is used to evaluate $\mathcal{L}_{\text{SDS}}$ which is optimized jointly with $\mathcal{L}_h$ for $N$ iterations. The combination of geometric and plausibility constraints improves the visual plausibility of the output while encouraging it to conform to the given constraints.

We note that the iterative approach in the \FirstStage{} leads to better results than alternative update strategies such as deforming the source mesh $\mathcal{M}_0$ by minimizing ARAP energy~\cite{Sorkine:2007ARAP} or, solving \eqn{}~\ref{eqn:poisson_constrained} using both $\mathbf{K}_h$ and $\mathbf{K}_a$ as hard constraints.
In our experiments (\shortsec{}~\ref{sec:experiments}), we show that both methods produce distortions that cannot be corrected by the diffusion prior in the subsequent stage. Specifically, directly solving \eqn{}~\ref{eqn:poisson_constrained} using all available constraints only yields the least squares solution $\mathbf{V}^*$ without updating the underlying Jacobians $\mathbf{J}$, resulting in the aforementioned distortions. %

\section{Experiments}
\label{sec:experiments}

We evaluate \APAP{} in downstream applications involving manipulation of 3D and 2D meshes.

\subsection{Experiment Setup}
\paragraph{Benchmark.}
To evaluate the plausibility of a mesh deformation we propose a novel benchmark \benchmark{} of textured 3D and 2D triangular meshes spanning both human-made and organic objects annotated with handle vertices and their editing directions, and anchor vertices.
The set of 3D meshes, \benchmarkThreeD{}, is constructed using meshes from ShapeNet~\cite{Chang:2015Shapenet} and \textit{Genie}~\cite{Luma:Genie}. The meshes are normalized to fit in a unit cube. Each mesh is manually annotated with editing instructions, including a set of anchors, handles, and corresponding targets to simulate editing scenarios.
\benchmark{} offers another subset called \benchmarkTwoD{}, a collection of 80 textured, planar meshes of various objects, to facilitate quantitative analysis and user study described later in this section.
To create \benchmarkTwoD{}, we first generate 2 images of real-world objects for each of the 20 categories using Stable Diffusion-XL~\cite{Podell:2023SDXL}.
We then extract foreground masks from the generated images using SAM~\cite{Kirillov:2023SAM} and sample pixels that lie on the boundary and interior. The sampled pixels are used for Delaunay triangulation,  constrained with the edges along the main contour of the masks, that produces 2D triangular meshes with texture.
We assign two handle and anchor pairs to each mesh that imitate user instructions.
For evaluation purposes, we populate the reference set by sampling $1,000$ images for each object category using Stable Diffusion-XL. The generated images are used to evaluate a perceptual metric to assess the plausibility of 2D mesh editing results as described in \shortsec{}~\ref{subsec:exp_image_editing}.

\begin{figure*}[!ht] 
{
\captionsetup{type=figure, labelfont=bf}
\centering
\setlength{\tabcolsep}{0em}
\def\arraystretch{0.0}
{\scriptsize
\begin{tabularx}{\textwidth}{m{0.11\textwidth}m{0.11\textwidth}m{0.08\textwidth}m{0.14\textwidth}m{0.08\textwidth}m{0.14\textwidth}m{0.09\textwidth}m{0.135\textwidth}m{0.115\textwidth}}
\multicolumn{3}{c}{View 1} & \multicolumn{3}{c}{View 2} & \multicolumn{3}{c}{View 2 (Zoom In)} \\
\makecell{Source} & \makecell{ARAP~\cite{Sorkine:2007ARAP}} & \makecell{Ours} & \makecell{Source} & \makecell{ARAP~\cite{Sorkine:2007ARAP}} & \makecell{Ours} & \makecell{Source} & \makecell{ARAP~\cite{Sorkine:2007ARAP}} & \makecell{Ours} \\
\multicolumn{9}{c}{
    \includegraphics[width=\textwidth]{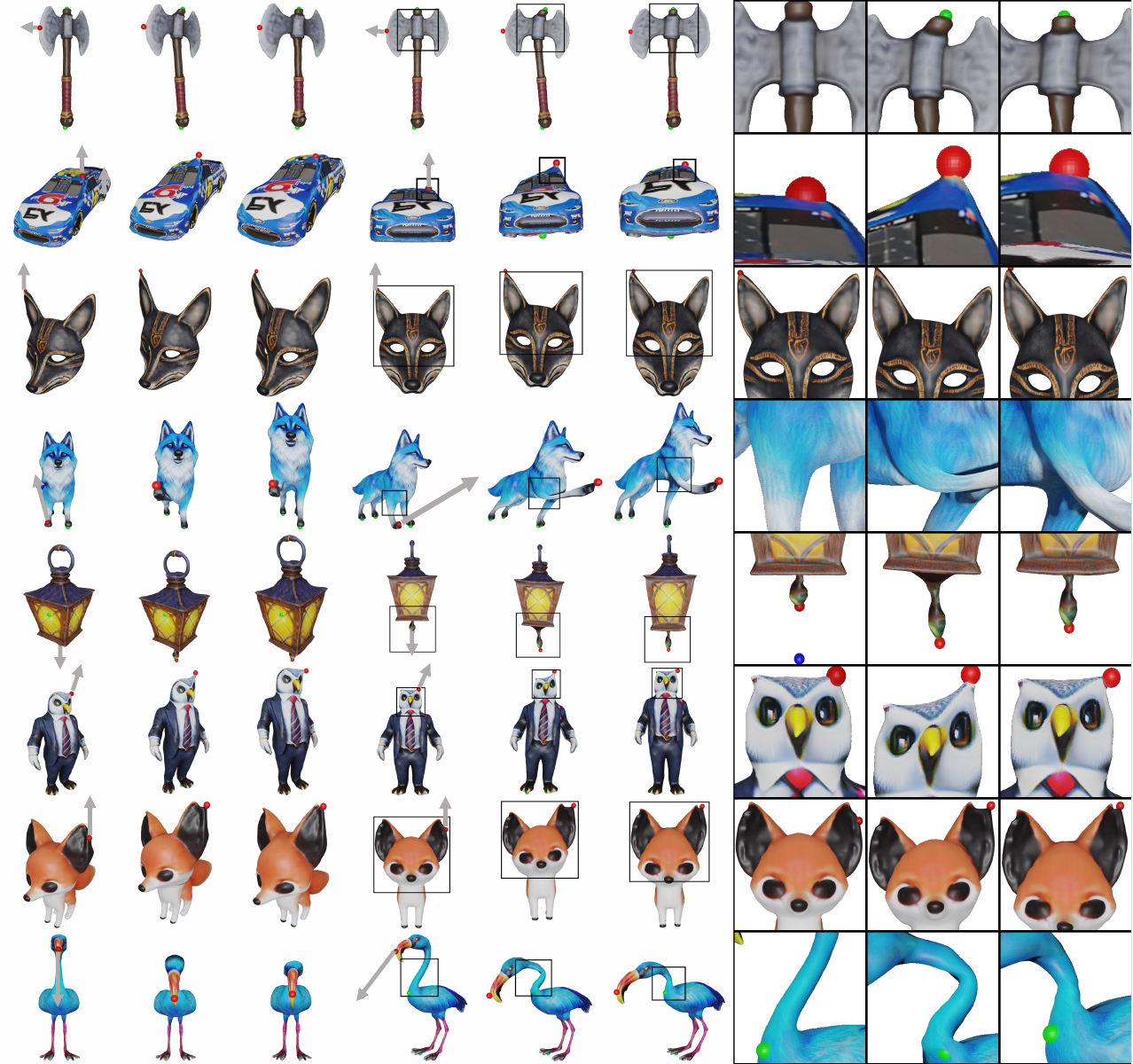}
}

\end{tabularx}
\vspace{-10pt}
\caption{
\textbf{Qualitative results from 3D shape deformation.} 
We visualize the source shapes and their deformations made using ARAP~\cite{Sorkine:2007ARAP} and ours by following the instructions each of which specifies a handle (\Red{red}), an edit direction denoted with an arrow (\Gray{gray}), and an anchor (\Green{green}). We showcase the rendered images captured from two different viewpoints, as well as one zoom-in view highlighting local details.
}
\label{fig:3d_result}
}
}
\end{figure*}

\paragraph{Baselines.}
We compare our method (\APAP{}) and As-Rigid-As-Possible (ARAP)~\cite{Sorkine:2007ARAP} since it is one of the widely used mesh deformation techniques that permits shape manipulation via direct vertex displacement.
Throughout the experiments, we use the implementation in \texttt{libigl}~\cite{Jacobson:2018libigl} with default parameters.

\vspace{-\baselineskip}
\paragraph{Evaluation Metrics.}
In 2D experiments, we conduct quantitative analysis based on $k$-NN GIQA score~\cite{Gu:2020GIQA} as an evaluation metric to assess the plausibility of instance-specific editing results. The metric quantifies the perceptual proximity between the edited image and its $k$ nearest neighbors in the reference set included in \benchmarkTwoD{}. As our objective is to make plausible variations of 2D meshes via deformation, an edited object should remain perceptually similar to other objects in the same category. We use $k=12$ throughout the experiments.

\subsection{3D Shape Deformation}
\label{subsec:exp_shape_deformation}

\paragraph{Qualitative Results.}
We showcase examples of 3D shape deformation where each deformation is specified by a handle (\Red{red}), an edit direction (\Gray{gray}), and an anchor (\Green{green}).
As shown in \fig~\ref{fig:3d_result}, \APAP{} is capable of manipulating 3D shapes to improve visual plausibility which is not achievable by solely relying on geometric prior such as ARAP~\cite{Sorkine:2007ARAP}.
For instance, given a user input that drags a handle on one blade of an axe (the first row) along an arrow, \APAP{} simultaneously expands both blades of the axe whereas ARAP~\cite{Sorkine:2007ARAP} produces distortions near the head. Similar examples that demonstrate symmetry-awareness of \APAP{} can be found in other cases such as a car (the second row), and an owl (the sixth row) where a user lifts only one side of the shape upward and the symmetry is recovered by \APAP{} which cannot be achieved by ARAP~\cite{Sorkine:2007ARAP}.
Also, note that \APAP{} is capable of making a smooth articulation at the leg of the wolf (the fourth row) by adjusting the overall posture in comparison to ARAP which creates an excess bending.

\begin{figure}[!t]
{
\captionsetup{type=figure, labelfont=bf}
\centering
\setlength{\tabcolsep}{0em}
\def\arraystretch{0.0}

{\scriptsize
\includegraphics[width=0.5\textwidth]{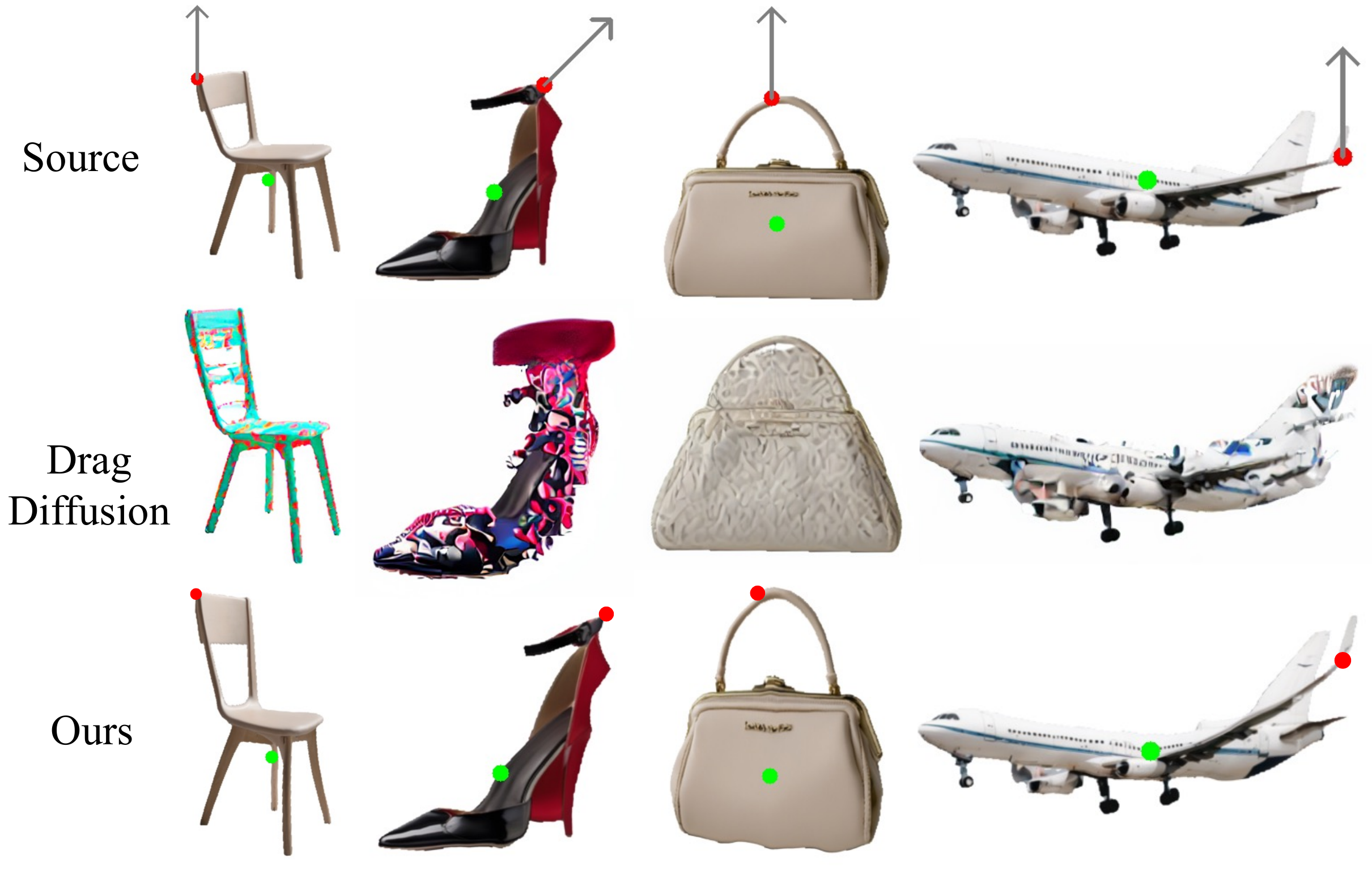} \\
}
\vspace{-5pt}
\caption{
    \textbf{Failure cases of DragDiffusion.}
    DragDiffusion~\cite{Shi:2023DragDiffusion} can easily compromise the identity of edited instances as it manipulates their latents without an explicit parameterization, the identity of instances can be broken during editing.
}
\label{fig:dragdiffusion_fail}
}
\end{figure}

\subsection{2D Mesh Editing}
\label{subsec:exp_image_editing}

\begin{figure*}[!t]
{
\captionsetup{type=figure, labelfont=bf}

\centering
\setlength{\tabcolsep}{0em}
\def\arraystretch{0.0}
{\scriptsize
\includegraphics[width=\textwidth]{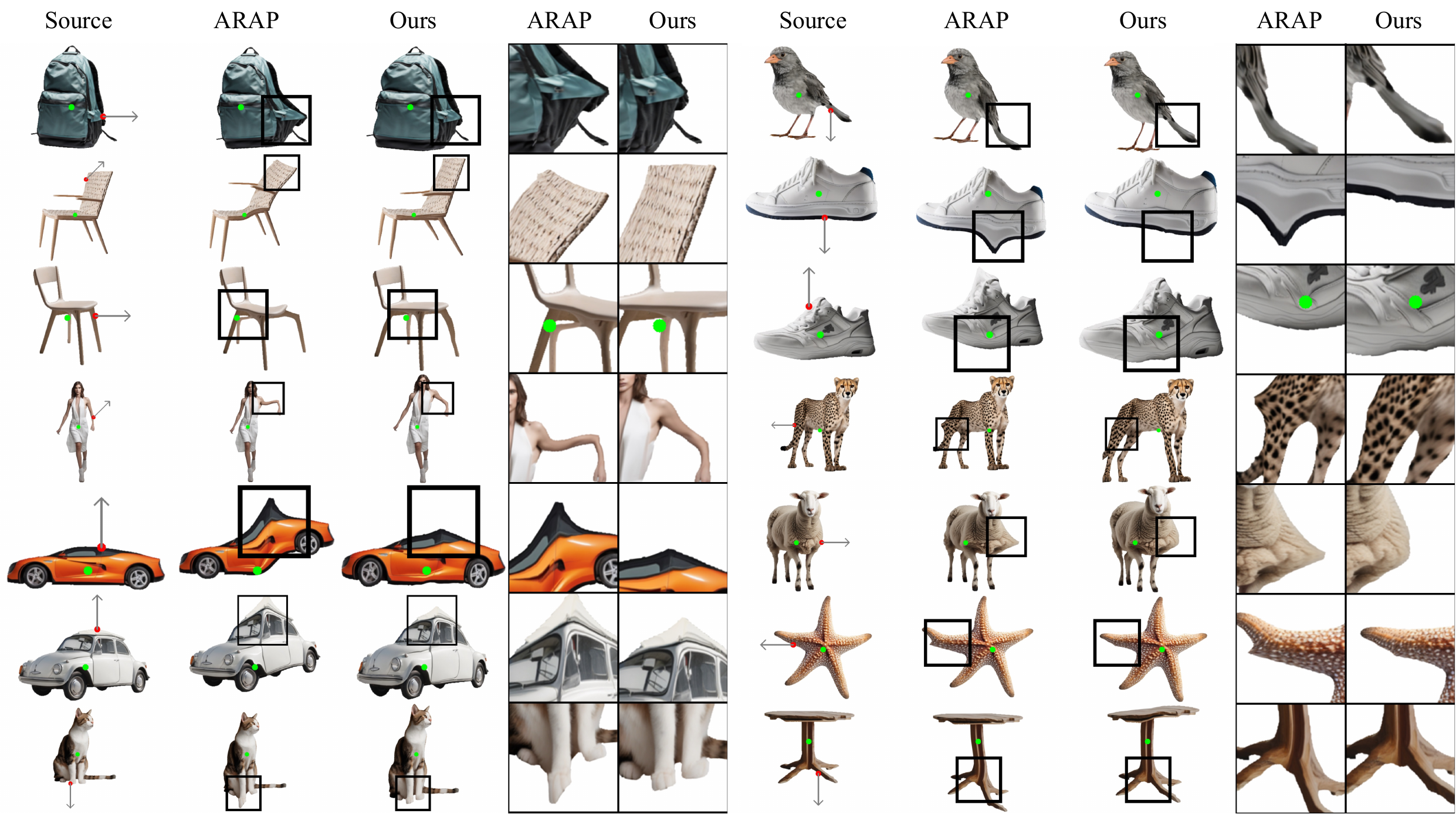}
\vspace{-10pt}
\caption{
\textbf{Qualitative results from 2D mesh deformation.} 
2D meshes are edited using ARAP~\cite{Sorkine:2007ARAP} and the proposed method following the edit instruction consisting of a handle (\Red{red}), a target direction (\Gray{gray}), and an anchor (\Green{green}).
We showcase the rendered images of the edited meshes, as well as a zoom-in view highlighting local details.
}
\label{fig:2d_result}
}
}
\end{figure*}

\paragraph{Qualitative Evaluation.}
We present qualitative results using the baselines and our method in \fig~\ref{fig:2d_result}. Each row shows two different results obtained by editing an image based on a handle moved from the original position (\Red{red}) along a direction indicated by an arrow (\Gray{gray}) while fixing an anchor (\Green{green}), similar to the 3D experiments discussed in the previous section. %

As shown in \fig~\ref{fig:2d_result}, ARAP~\cite{Sorkine:2007ARAP} enforces local rigidity and often results in implausible deformations. For example, it does not account for the mechanics of the human body and introduces an unrealistic articulation of a human arm (the fourth row). In addition, it twists the body of a sports car (the fifth row). Both of them originate from the lack of understanding of the appearance of objects. \APAP{} alleviates this issue by incorporating a visual prior into shape deformation producing a bending near the elbow and preserving the smooth silhouette of the car, respectively.

While \APAP{} is designed for meshes not images, we provide an additional qualitative comparison against DragDiffusion~\cite{Shi:2023DragDiffusion}, an image editing technique that operates in pixel space, to demonstrate the effectiveness of mesh-based parameterization in applications where identity preservation is crucial. As shown in \fig{}~\ref{fig:dragdiffusion_fail}, DragDiffusion~\cite{Shi:2023DragDiffusion} may corrupt the identity of the instances depicted in input images during the encoding and decoding procedure. \APAP{}, on the other hand, makes plausible variations of the given objects while maintaining their originality, benefiting from an explicit mesh representation it is grounded.

\begin{table}[t!]
    \centering
{
\footnotesize
\setlength{\tabcolsep}{0.2em}
\begin{tabularx}{\linewidth}{>{\centering}m{2.3cm}| Y}
\toprule
Methods & $k$-NN GIQA ($\times 10^{-2}$) $\uparrow$ \\
\midrule 
ARAP~\cite{Sorkine:2007ARAP}               & 4.753          \\
DragDiffusion~\cite{Shi:2023DragDiffusion} & 4.545          \\
Ours ($\mathcal{L}_h$ Only)                & 4.797          \\
Ours (ARAP Init.)                          & 4.740          \\
Ours (Poisson Init.)                       & 4.316          \\
Ours                                       & \textbf{4.887} \\
\bottomrule
\end{tabularx}
}
\vspace{-5pt}
\caption{
\textbf{Quantitative analysis for 2D mesh editing.}
\projname{} outperforms its baselines in quantitative evaluation using $k$-NN GIQA~\cite{Gu:2020GIQA}.
}
\label{tbl:giqa}
\end{table}

\begin{table}[t!]
    \centering
{
\footnotesize
\setlength{\tabcolsep}{0em}
\renewcommand{\arraystretch}{1.0}
\begin{tabularx}{\linewidth}{>{\centering}m{2.3cm}| YY}
\toprule
Methods & Preference (\%) $\uparrow$\\
\midrule 
ARAP~\cite{Sorkine:2007ARAP}               & 40.83 \\
Ours                                       & \textbf{59.17}\\
\bottomrule
\end{tabularx}
}
\vspace{-5pt}
\caption{
\textbf{User study preference for 2D image editing.}
In a user study targeting users on Amazon Mechanical Turk (MTurk), the results produced using ours were preferred over the outputs from the baseline.
}
\label{tbl:user_study}
\end{table}

\vspace{-\baselineskip}
\paragraph{Quantitative Evaluation.}
\tab{}~\ref{tbl:giqa} summarizes $k$-NN GIQA scores measured on the outputs from ARAP~\cite{Sorkine:2007ARAP} (the first row) and \APAP{} (the sixth row) using \benchmarkTwoD{}. As shown, \APAP{} demonstrates superior performance over ARAP~\cite{Sorkine:2007ARAP}. This again verifies the observations from qualitative evaluation where ARAP~\cite{Sorkine:2007ARAP} introduces distortions that harm visual plausibility. As in qualitative evaluation, we also report the $k$-NN GIQA score of DragDiffusion~\cite{Shi:2023DragDiffusion}, degraded due to artifacts caused during direct manipulation of latents.

\vspace{-\baselineskip}
\paragraph{User Study.}
We further conduct a user study for a more precise perceptual analysis. We follow Ritchie~\cite{Ritchie:MTurk} and recruit participants on Amazon Mechanical Turk (MTurk). Each participant is provided with a set of $20$ randomly sampled images of the source meshes paired with editing results of ARAP~\cite{Sorkine:2007ARAP} and \APAP{}. 
To check whether the response from a participant is reliable we present $5$ vigilance tests and collect $102$ responses from the participants who passed the vigilance test.

We instructed participants to select the most anticipated outcome when the displayed source image is edited by the dragging operation visualized as an arrow.
We provide details of the user study in \suppl{}.
\tab~\ref{tbl:user_study} shows a higher preference of the participants on our method over ARAP~\cite{Sorkine:2007ARAP} implying that our method produces more visually plausible deformations.

\vspace{-\baselineskip}
\paragraph{Ablation Study.}
\tab{}~\ref{tbl:giqa} summarizes the impact of different initialization strategies in the first stage on $k$-NN GIQA score. As reported in the third row of the table, optimizing $\mathcal{L}_h$ that aims to exclusively satisfy geometric constraints leads to unnatural distortions. We provide a qualitative comparison in the \suppl{}.

While designing the algorithm illustrated in \alg~\ref{alg:apap_main_loop}, we considered other options for \texttt{FirstStage}. Instead of optimizing $\mathcal{L}_h$ to initially deform a shape, we used a shape produced by ARAP~\cite{Sorkine:2007ARAP} or by solving a Poisson's equation constrained not only on anchor positions but also on handles at their target positions reached by following the given edit directions. We report $k$-NN GIQA scores of the alternatives in the fourth and fifth row of \tab~\ref{tbl:giqa}, respectively. Both initialization strategies degrade the plausibility of results due to large distortions introduced by either solely enforcing local rigidity or, finding least square solutions without updating Jacobians. This poses a challenge to the diffusion prior, making it struggle to induce meaningful update directions when provided with renderings with noticeable distortions, which can be found in qualitative analysis in \suppl{}.

\section{Conclusion}
\label{sec:conclusion}
We presented \APAP{}, a novel deformation framework that tackles the problem of plausibility-aware shape deformation while offering intuitive controls over a wide range of shapes represented as triangular meshes. To this end, we carefully orchestrate two core components, a learnable Jacobian-based parameterization that originates from geometry processing and powerful 2D priors acquired by text-to-image diffusion models trained on Internet-scale datasets. We assessed the performance of the proposed method against an existing geometric-prior-based deformation technique and also thoroughly investigated the significance of our design choices through experiments.

\vspace{-\baselineskip}
\paragraph{Acknowledgements}
We appreciate RECON Labs for their support in our experiments.
This work was supported by
NRF grant (RS-2023-00209723) and IITP grants (2022-0-
00594, RS-2023-00227592) funded by the Korean government (MSIT), Seoul R\&BD Program (CY230112), and
grants from the DRB-KAIST SketchTheFuture Research Center, Hyundai NGV, KT, NCSOFT, and Samsung Electronics.

{\small
\bibliographystyle{ieee_fullname}
\bibliography{egbib}
}

\newif\ifpaper
\papertrue

\renewcommand{\thesection}{A}
\renewcommand{\thetable}{A\arabic{table}}
\renewcommand{\thefigure}{A\arabic{figure}}

\clearpage
\newpage
\onecolumn
\section*{Appendix}
\label{sec:appendix}
\ifpaper
    \newcommand{\refofpaper}[1]{\unskip}
    \newcommand{\refinpaper}[1]{\unskip}
\else
    \makeatletter
    \newcommand{\manuallabel}[2]{\def\@currentlabel{#2}\label{#1}}
    \makeatother

    \manuallabel{sec:intro}{Sec. 1}
    \manuallabel{sec:related_work}{Sec. 2}
    \manuallabel{sec:method}{Sec. 3}
    \manuallabel{subsec:deform_param}{Sec. 3.1}
    \manuallabel{subsec:score_distillation}{Sec. 3.2}
    \manuallabel{subsec:as_plausible_as_possible}{Sec. 3.3}
    \manuallabel{sec:experiments}{Sec. 4}

    \manuallabel{eq:biharmonics_solution}{Eqn. 2}
    \manuallabel{eq:tattas}{Eqn. 6}
    \manuallabel{eq:our_reformulation}{Eqn. 11}
    
    \manuallabel{fig:teaser}{Fig. 1}
    \manuallabel{fig:method}{Fig. 2}
    \manuallabel{fig:2d_result}{Fig. 3}

    \manuallabel{alg:apap_main_loop}{1}

    \manuallabel{tbl:giqa}{Tab. 1}
    \manuallabel{tbl:l2}{Tab. 2}
    \manuallabel{tbl:user_study}{Tab. 3}
    
    \newcommand{\refofpaper}[1]{of the main paper}
    \newcommand{\refinpaper}[1]{in the main paper}
\fi
In this supplementary material, we first describe implementation details of our main pipeline (\shortsec{}~\ref{sec:implementation_details}) and details of \benchmark{} construction (\shortsec{}~\ref{sec:bench_mark_detail}). We also provide the exact question given to user study participants, as well as an example questionnaire (\shortsec{}~\ref{sec:user_study_detail}). Furthermore, we summarize additional experimental results including an ablation study for 2D mesh editing based on qualitative results (\shortsec{}~\ref{sec:ablation_2d}), additional qualitative results for 3D shape deformation (\shortsec{}~\ref{sec:ablation_study_3d}), more complex 3D shape deformations achieved by leveraging classical deformation techniques (\shortsec{}~\ref{sec:complex_deform_3d}), and human evaluation on the plausibility of 3D deformations via user study (\shortsec{}~\ref{sec:user_study_3d}).

\subsection{Implementation Details}
\label{sec:implementation_details}
We provide additional implementation details of \alg{}~\ref{alg:apap_main_loop}~\refinpaper{}.
We used a modified version of the differentiable Poisson solver from \cite{Aigerman:2022NJF}, denoted by $g$ in \alg{}~\ref{alg:apap_main_loop}, and \texttt{nvdiffrast}~\cite{Laine:2020nvdiffrast} when implementing the differentiable renderer $\mathcal{R}$ in our pipeline. We render 2D/3D meshes at a resolution of $512 \times 512$.

When editing 2D meshes, we optimize $\mathcal{L}_h$ for $M=300$ iterations in the \FirstStage{} and jointly optimize $\mathcal{L}_h$ and $\mathcal{L}_{\text{SDS}}$ for $N=700$ iterations in the \SecondStage{}.
For experiments involving the optimization of 3D meshes with increased geometric complexity, we use $M=300$ and $N=1000$ for each stage, respectively. We use ADAM~\cite{Kingma:2015Adam} with a learning rate $\gamma=1 \times 10^{-3}$ throughout the optimization. We use the Classifier-Free Guidance (CFG) scale of 100.0 and randomly sample $t \in [0.02, 0.98]$ when evaluating $\mathcal{L}_{\text{SDS}}$ following DreamFusion~\cite{Poole:2023Dreamfusion}.

We use a script from \textit{diffusers}~\cite{HF:diffusers} to finetune Stable Diffusion~\cite{Rombach:2022LDM} with LoRA~\cite{Hu:2022LoRA}. We employ \texttt{stabilityai/stable-diffusion-2-1-base} as our base model and augment its cross-attention layers in the U-Net with rank decomposition matrices of rank $16$.
For the task of 2D mesh editing, we train the injected parameters for $60$ iterations, utilizing a rendering of a mesh as a training image. In the 3D shape deformation, where renderings from 4 canonical viewpoints (front, back, left, and right) are available, we finetune the model for $200$ iterations. In both cases, we use the learning rate $\gamma = 5 \times 10^{-4}$.

\subsection{Details of \benchmark{}}
\label{sec:bench_mark_detail}
\paragraph{Image Generation.} For evaluation purposes, we build \benchmarkTwoD{} by generating 2 images of real-world objects for each of the 20 categories using Stable Diffusion-XL~\cite{Podell:2023SDXL} as noted in the main paper. We segment the foreground objects from the generated images and run Delaunay triangulation to populate a collection of 2D meshes.
When generating the images, we use the following template prompt~\texttt{"a photo of [category name] in a white background"} for all categories to facilitate foreground object segmentation. \tab{}~\ref{tbl:suppl_categories} summarizes the list of categories. Note that the list includes both human-made and organic objects that can be easily found in the daily environment to test the generalization capability of a deformation technique to various object types.

\begin{table}[h!]
    \centering
{
\footnotesize
\setlength{\tabcolsep}{0.2em}
\begin{tabularx}{\linewidth}{Y|Y}
\toprule
Human-Made & Organic \\
\midrule
backpack & flying bird \\
bike & side view of cat \\
chair & side view of dog \\
high-heeled shoes & runway model  \\
purse & sitting bird \\
side view of car & standing cheetah \\
sneakers & standing dragon \\
table & standing raccoon \\
airplane & standing sheep \\
& standing white duck \\
& starfish \\
\bottomrule
\end{tabularx}
}
\caption{\textbf{Object categories of 2D meshes in \benchmarkTwoD{}.} \benchmarkTwoD{} includes 2D triangle meshes depicting various objects, including both human-made and organic objects.}
\label{tbl:suppl_categories}
\end{table}

\vspace{-\baselineskip}
\paragraph{Handle and Anchor Assignment.}
We manually assign two handle and anchor pairs to each mesh to imitate user instructions. Specifically, we choose vertices on the shape boundaries instead of internal vertices to induce deformations that alter object silhouettes. For instance, users would try to drag the bottom of a backpack downward to enlarge the shape, instead of dragging an interior point which may flip triangles, distorting the appearance. As an anchor, we use the vertex closest to the center of mass of each mesh.

In experiments using \benchmarkThreeD{} and \benchmarkTwoD{}, we note that utilization of neighboring vertices of the given handles and anchors during deformation helps retain smooth geometry near the handle. Therefore, we additionally sample vertices near the handles and anchors that lie in the sphere of radius $r=0.01$ and denote the extended sets of handles and anchors \textit{region handles} and \textit{region anchors}, respectively. We use region anchors and a single handle for 3D experiments and region anchors and region handles for 2D cases. Note that we use the same sets of handles and anchors when deforming shapes with our baselines for fair comparisons.

\subsection{Details of User Study for 2D Mesh Editing}
\label{sec:user_study_detail}

In \shortsec{}~\ref{sec:experiments} of the main paper, we reported the preference statistics collected from 102 user study participants who passed the vigilance tests. We provide additional details of the user study in the following. We instructed participants to select the most anticipated outcome when the displayed source image is edited by the dragging operation visualized as an arrow with the question: \texttt{"A visual designer wants to modify the object by clicking on a red point and dragging it in the direction of the arrow. Please choose a result that best satisfies the designer's edit, while retaining the characteristics and plausibility of the object."}

\begin{figure*}[h!]
    \captionsetup{type=figure, labelfont=bf}
    \centering
    \footnotesize{
        \renewcommand{\arraystretch}{0.0}
        \setlength{\tabcolsep}{0.0em}
        \setlength{\fboxrule}{0.0pt}
        \setlength{\fboxsep}{0pt}
        \begin{tabularx}{\textwidth}{>{\centering\arraybackslash}m{0.5\textwidth} >{\centering\arraybackslash}m{0.5\textwidth}}
        \includegraphics[width=0.49\textwidth]{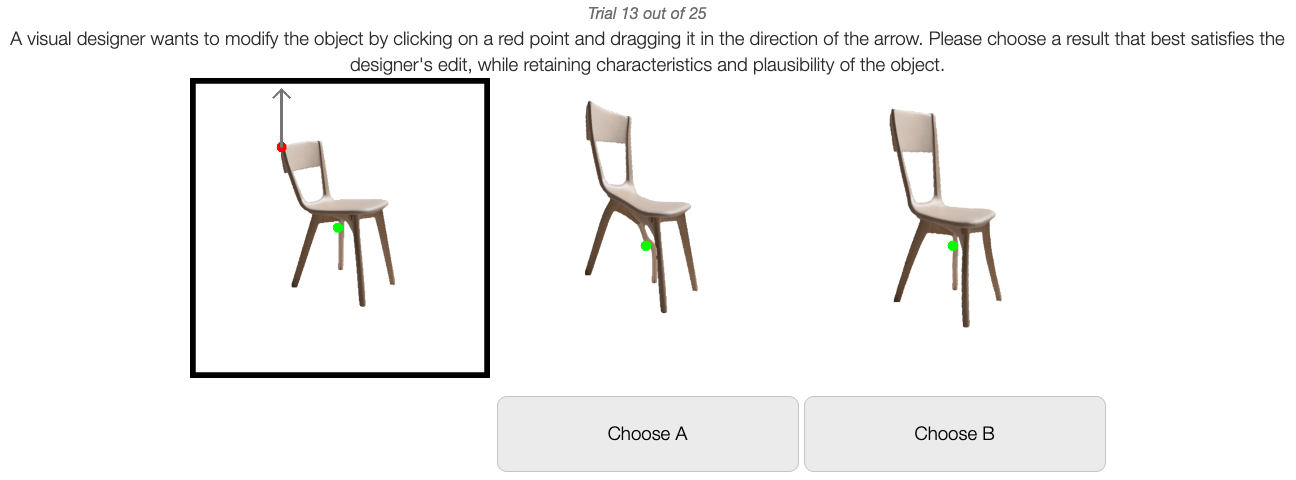} &
        \includegraphics[width=0.49\textwidth]{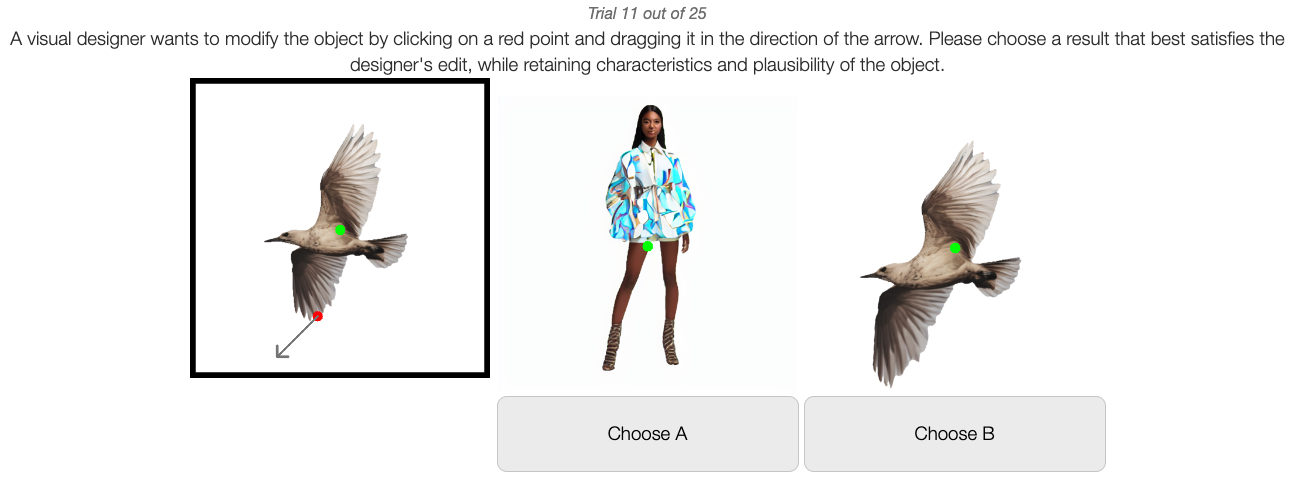}
        \end{tabularx}
    }
    \caption{\textbf{Examples of questionnaires displayed during the user study (2D mesh editing).} During the user study, we asked the participants to evaluate 20 different result pairs from ARAP~\cite{Sorkine:2007ARAP} and ours as shown on the left. To check whether a participant is focusing on the user study, we included 5 items for the vigilance test. As shown on the right, a question for the vigilance test includes an image of an object that is not related to the source image.}
    \label{fig:userstudy}
\end{figure*}

\fig{}~\ref{fig:userstudy} (left) shows an example of a questionnaire provided to the participants. For vigilance tests, we included an editing result from DragDiffusion~\cite{Shi:2023DragDiffusion} depicting an object irrelevant to the source image in each question. The participants were asked to answer the same question. We illustrate an example questionnaire of a vigilance test in \fig{}~\ref{fig:userstudy} (right).

\vspace{-1\baselineskip}
\subsection{Additional Qualitative Results for 3D Shape Deformation}
\label{sec:ablation_study_3d}

\begin{figure*}[t!] 
{
\captionsetup{type=figure, labelfont=bf}
\centering
\setlength{\tabcolsep}{0em}
\def\arraystretch{0.0}
{\scriptsize
\begin{tabularx}{\textwidth}
{m{0.08\textwidth}m{0.11\textwidth}m{0.10\textwidth}m{0.13\textwidth}m{0.12\textwidth}m{0.12\textwidth}m{0.115\textwidth}m{0.110\textwidth}m{0.115\textwidth}}
\multicolumn{3}{c}{View 1} & \multicolumn{3}{c}{View 2} & \multicolumn{3}{c}{View 2 (Zoom In)} \\
\makecell{Source} & \makecell{ARAP~\cite{Sorkine:2007ARAP}} & \makecell{Ours} & \makecell{Source} & \makecell{ARAP~\cite{Sorkine:2007ARAP}} & \makecell{Ours} & \makecell{Source} & \makecell{ARAP~\cite{Sorkine:2007ARAP}} & \makecell{Ours} \\
\multicolumn{9}{c}{
    \includegraphics[width=\textwidth]{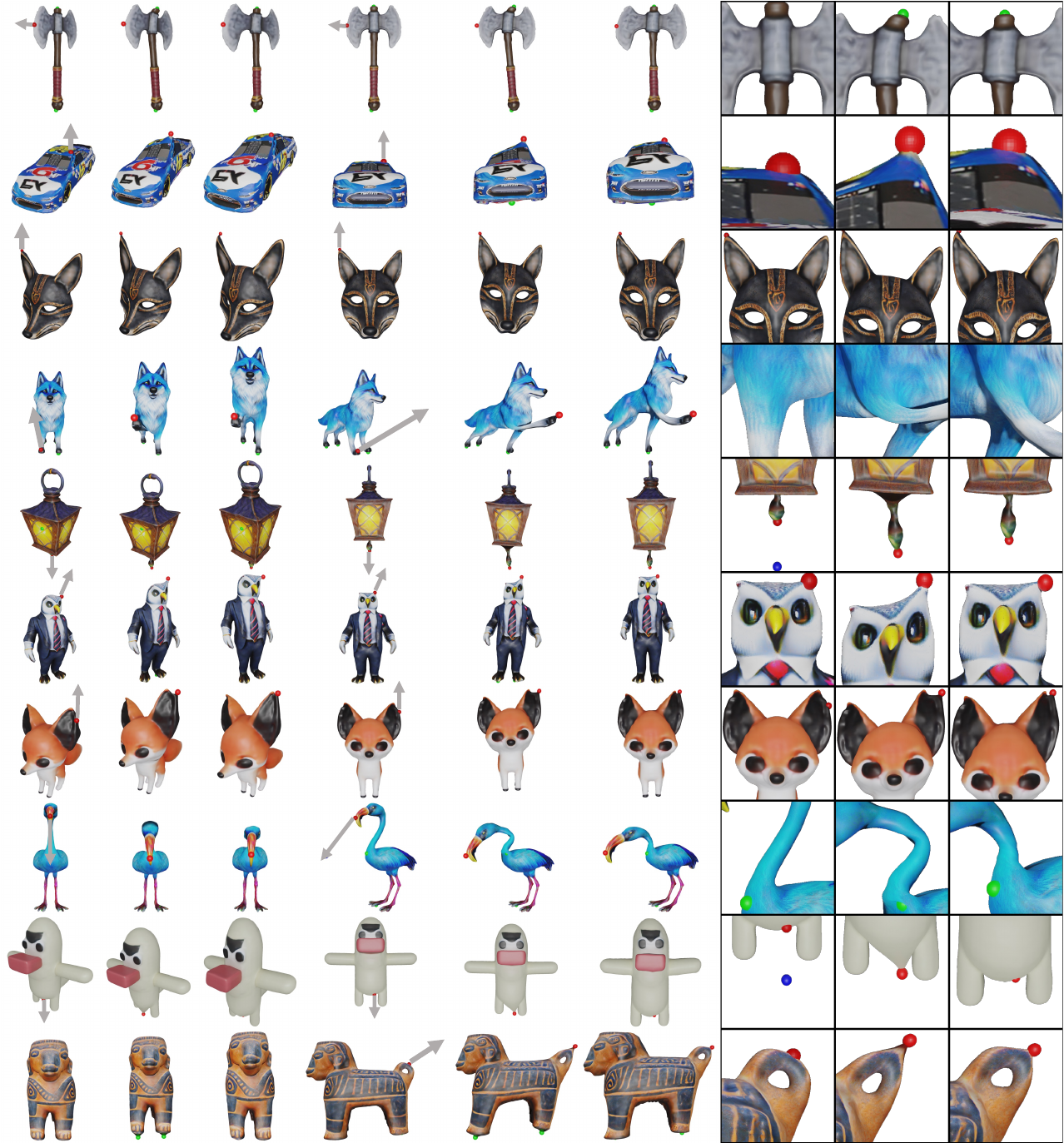}
}
\end{tabularx}
\vspace{-10pt}
\caption{
\textbf{Additional qualitative results from 3D shape deformation.} 
We visualize the source shapes and their deformations made using ARAP~\cite{Sorkine:2007ARAP} and ours by following the instructions each of which specifies a handle (\Red{red}), an edit direction denoted with an arrow (\Gray{gray}), and an anchor (\Green{green}). We showcase the rendered images captured from two different viewpoints, as well as one zoom-in view highlighting local details.
}
\label{fig:suppl_3d_result}
}
}
\end{figure*}

\fig{}~\ref{fig:suppl_3d_result} summarizes outputs of 3D shape deformation with additional results. As reported \refinpaper{}, ARAP~\cite{Sorkine:2007ARAP} only enforces local rigidity and hence cannot produce smooth deformations intended by users. In the ninth row, ARAP~\cite{Sorkine:2007ARAP} introduces a pointy end given an editing instruction that drags the bottom of a doll downward. Ours, however, elongates the entire geometry smoothly, producing a more visually plausible deformation. Another example displayed in the tenth row shows similar behaviors of ARAP~\cite{Sorkine:2007ARAP} and ours, respectively. Here, unlike ARAP~\cite{Sorkine:2007ARAP}, the proposed method adjusts the overall proportion of the statue as the handle located at the tail is translated, while preserving the smooth and round geometry near the handle.

\subsection{Complex 3D Deformation Examples}
\label{sec:complex_deform_3d}
In addition to the ability to optimize Jacobian fields using diffusion priors offered by the linearity of Poisson solvers, we can directly propagate local transforms, additionally defined at handle vertices, to Jacobians of neighboring faces by employing geodesic distances as weights~\cite{Yu:2004Poisson}. This allows for more dramatic deformations illustrated in \fig{}~\ref{fig:complex_3d_result}, involving limb articulations, large bending, and the use of multiple handles and anchors. As represented in the \textit{Panda} (the seventh, eighth, ninth columns) example, our framework can handle large pose variations, useful in downstream applications, such as animation.

\begin{center}
    \captionsetup{type=figure, labelfont=bf}
    \centering
    \setlength{\tabcolsep}{0em}
    \def\arraystretch{0.0}
    {\scriptsize
        \begin{tabularx}{\textwidth}{YYYYYYYYYYYY}
            \makecell{Source\\(View 1)} & \makecell{Ours\\(View 1)} & \makecell{Ours\\(View 2)} & \makecell{Source\\(View 1)} & \makecell{Ours\\(View 1)} & \makecell{Ours\\(View 2)} & \makecell{Source\\(View 1)} & \makecell{Ours\\(View 1)} & \makecell{Ours\\(View 2)} & \makecell{Source\\(View 1)} & \makecell{Ours\\(View 1)} & \makecell{Ours\\(View 2)} \\
            \includegraphics[width=\textwidth]{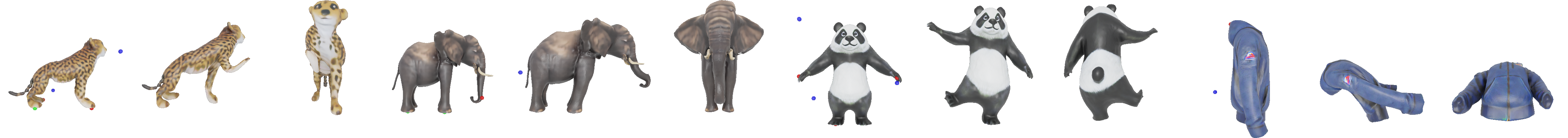}
        \end{tabularx}
    \caption{Examples of deforming source meshes using multiple handles and anchors. Best viewed in Zoom-in.}
    \label{fig:complex_3d_result}
    }    
\end{center}

\subsection{User Study Results for 3D Shape Deformation}
\label{sec:user_study_3d}

Assessing the visual plausibility of 3D deformations is particularly challenging due to the difficulty in populating large-scale reference sets as we did for 2D meshes in \shortsec{}~\ref{sec:experiments}~\refinpaper{}.
We further note that, unlike 3D generative models, computing image-based metrics such as CLIP-R score is non-trivial since it is hard to describe handle-based deformations solely using text prompts.

\begin{figure*}[h!]
    \captionsetup{type=figure, labelfont=bf}
    \centering
    \footnotesize{
        \renewcommand{\arraystretch}{0.0}
        \setlength{\tabcolsep}{0.0em}
        \setlength{\fboxrule}{0.0pt}
        \setlength{\fboxsep}{0pt}
        \begin{tabularx}{\textwidth}{>{\centering\arraybackslash}m{0.5\textwidth} >{\centering\arraybackslash}m{0.5\textwidth}}
        \includegraphics[width=0.49\textwidth]{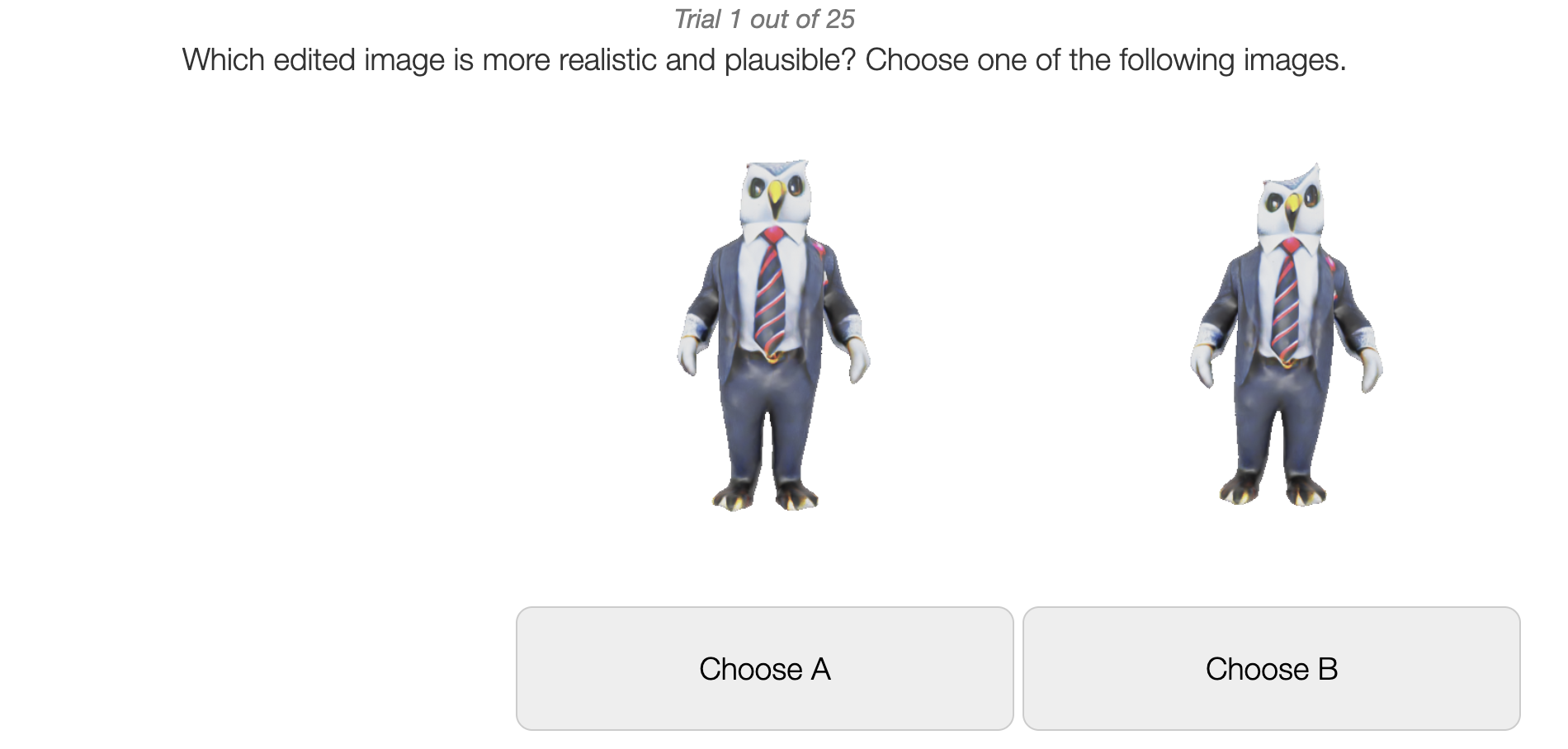} &
        \includegraphics[width=0.49\textwidth]{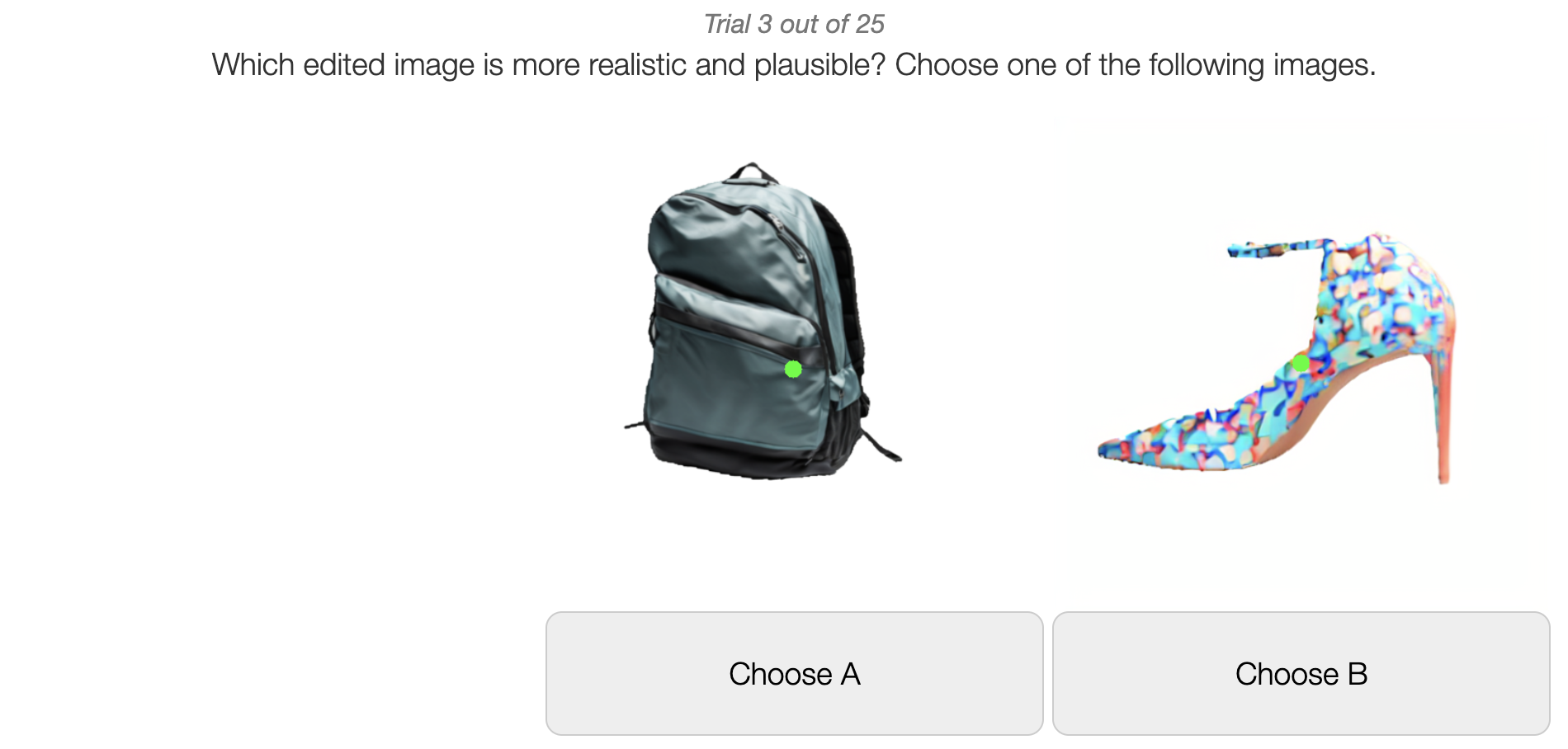}
        \end{tabularx}
    }
    \caption{\textbf{Examples of questionnaires displayed during the user study (3D shape deformation).} During the user study, we asked the participants to evaluate 20 different result pairs from ARAP~\cite{Sorkine:2007ARAP} and ours as shown on the left. To check whether a participant is focusing on the user study, we included 5 items for the vigilance test. As shown on the right, a vigilance test asks a participant to compare two images, with one of them containing noticeable artifacts.}
    \label{fig:userstudy_3d}
    \vspace{-\baselineskip}
\end{figure*}

Therefore, we conduct a user study similar to the one presented in \shortsec{}~\ref{sec:experiments}~\refinpaper{}.
We asked 47 user study participants on Amazon Mechanical Turk (MTurk) to compare rendered images of meshes deformed using ARAP~\cite{Sorkine:2007ARAP} and ours.
Each participant is provided with 20 image pairs and asked to select one image at each time given the question: \texttt{"Which edited image is more realistic and plausible? Choose one of the following images."}
An example of a questionnaire displayed to the participants is shown in \fig{}~\ref{fig:userstudy_3d} (left).
We provide an example of vigilance tests, similar to the user study for 2D mesh editing, in \fig{}~\ref{fig:userstudy_3d} (right).
As summarized in \tab{}~\ref{tbl:user_study_3d}, the deformation produced by our method is preferred over the results from the baseline.

\begin{table}[h!]
    \centering
{
\footnotesize
\setlength{\tabcolsep}{0em}
\renewcommand{\arraystretch}{1.0}
\begin{tabularx}{\linewidth}{>{\centering}m{2.3cm}| YY}
\toprule
Methods & Preference (\%) $\uparrow$\\
\midrule 
ARAP~\cite{Sorkine:2007ARAP}               & 41.7 \\
Ours                                       & \textbf{58.3}\\
\bottomrule
\end{tabularx}
}
\vspace{-5pt}
\caption{
\textbf{User study preference for 3D mesh deformation.}
In a user study targeting users on Amazon Mechanical Turk (MTurk), the results produced using ours were preferred over the outputs from the baseline.
}
\label{tbl:user_study_3d}
\end{table}

\subsection{Ablation Study for 2D Mesh Editing}
\label{sec:ablation_2d}
In this section, we provide qualitative results from the ablation study to validate the impact of each component on the plausibility of editing results. In \fig{}~\ref{fig:suppl_2d_ablation}, we summarize the results obtained by
(1) optimizing only $\mathcal{L}_h$, (2) $\mathcal{L}_h$ and $\mathcal{L}_{\text{SDS}}$ without LoRA finetuning, (3) skipping the \FirstStage{}, (4) using ARAP initialization, (5) using Poisson initialization, and (6) Ours.
As mentioned in the main paper, optimizing only $\mathcal{L}_h$ (the second column) either distorts texture (the fifth row) or inflates or shrinks other parts of the given shape (the seventh and twelfth row). This demonstrates the necessity of a visual prior during deformation.
Also, we observe the cases where skipping the \FirstStage{} (the fourth column) does not lead to intended deformation as our diffusion prior is reluctant to modify shapes from their original states (the first, second, and fifth row). 
On the other hand, deformations initialized with the meshes produced by ARAP~\cite{Sorkine:2007ARAP} (the fifth column) or Poisson solve (the sixth column) suffer from distortions that could not be resolved by optimizing $\mathcal{L}_{\text{SDS}}$ in the \SecondStage{}.

\newcommand{\AllComparisonImages}{\Image{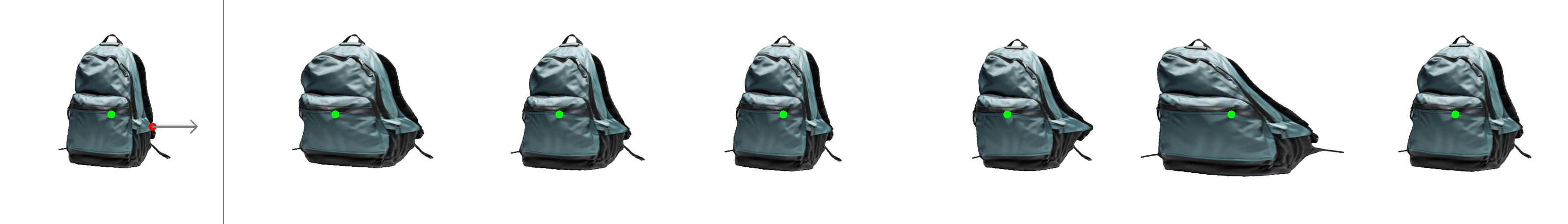} 
\Image{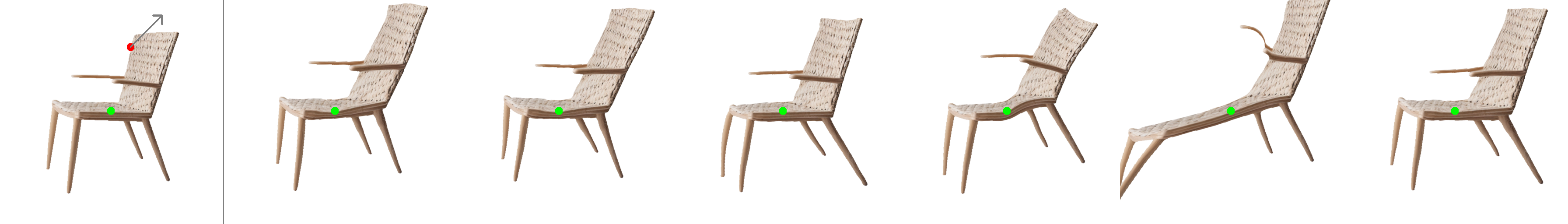} 
\Image{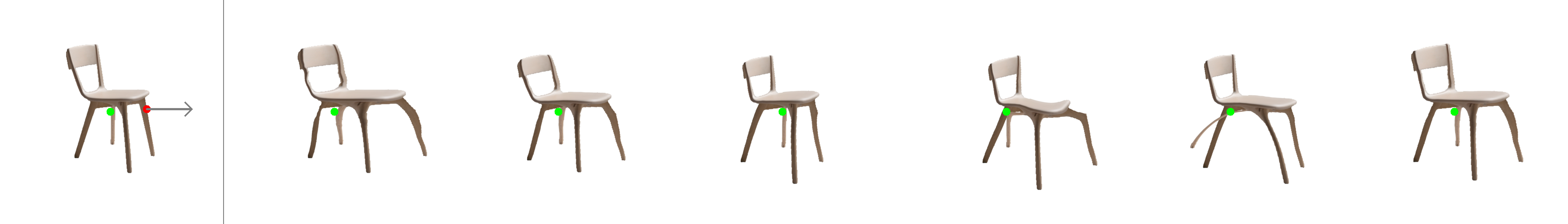} 
\Image{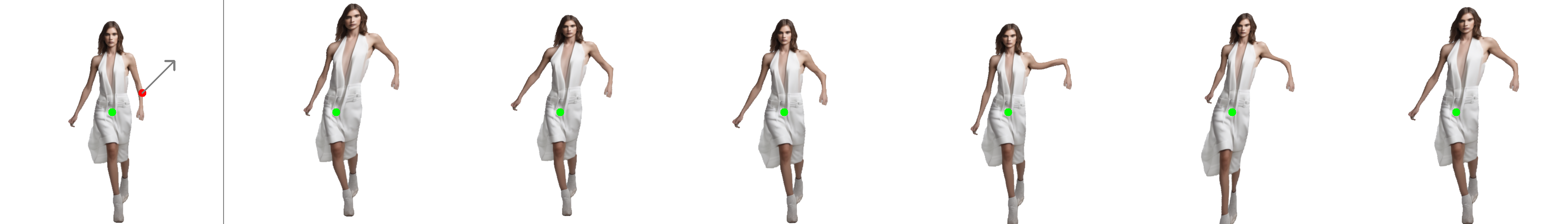} 
\Image{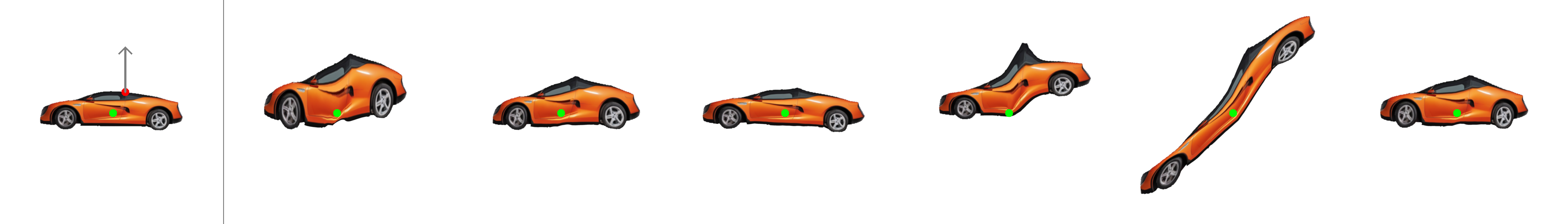} 
\Image{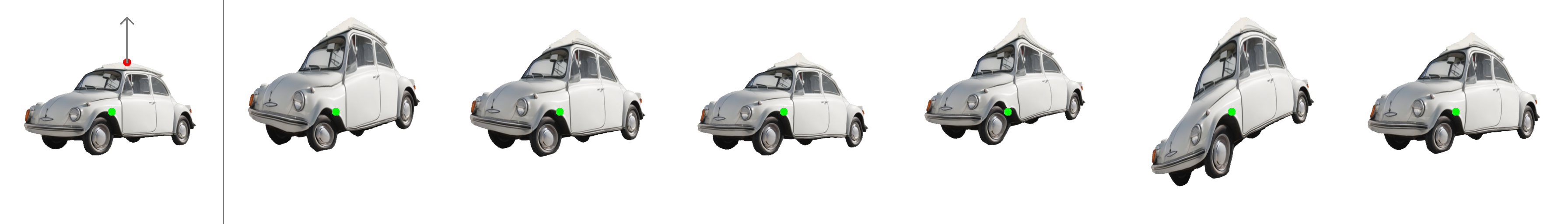} 
\Image{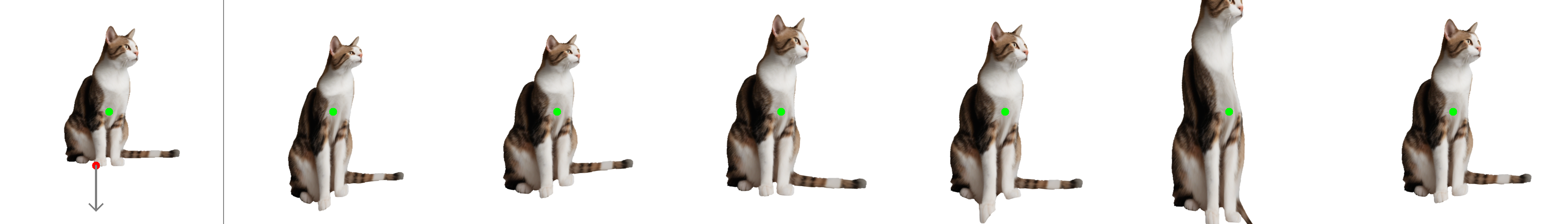} 
\Image{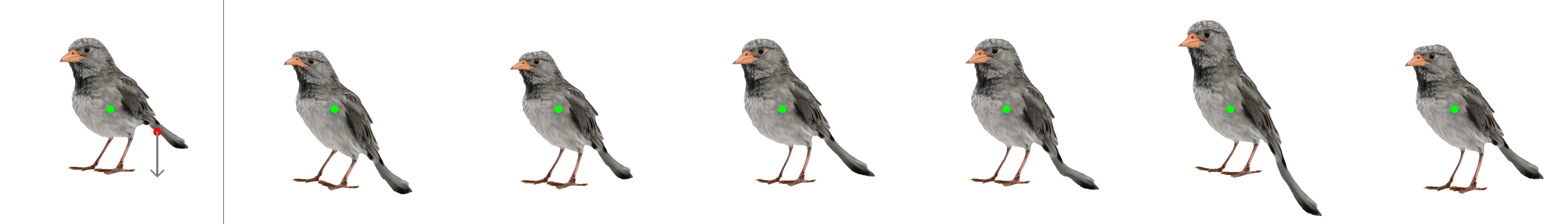} 
\Image{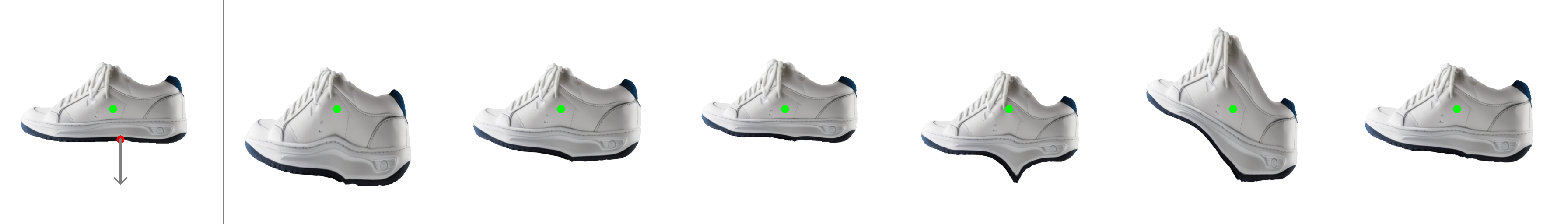} 
\Image{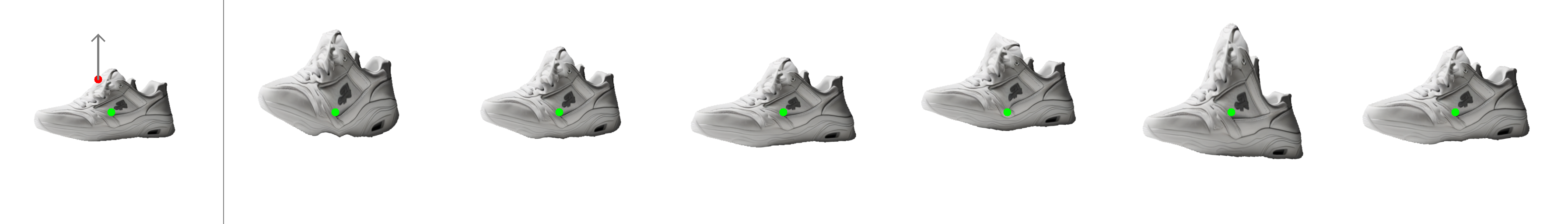} 
\Image{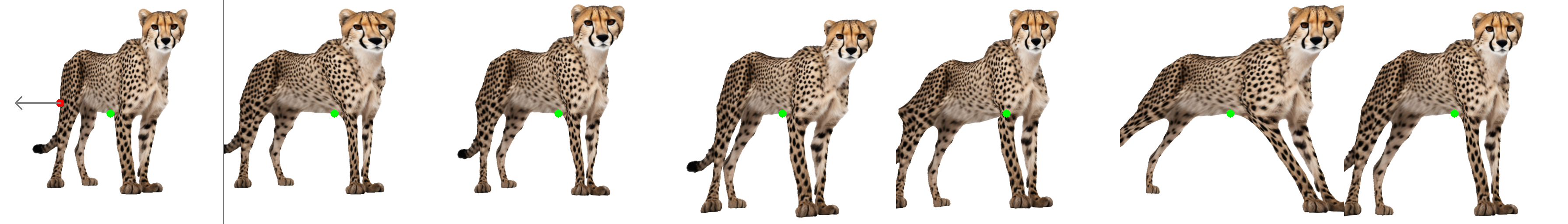} 
\Image{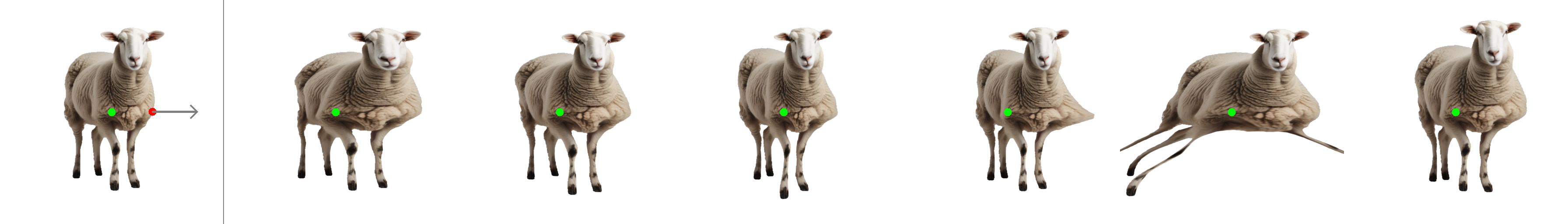} 
\Image{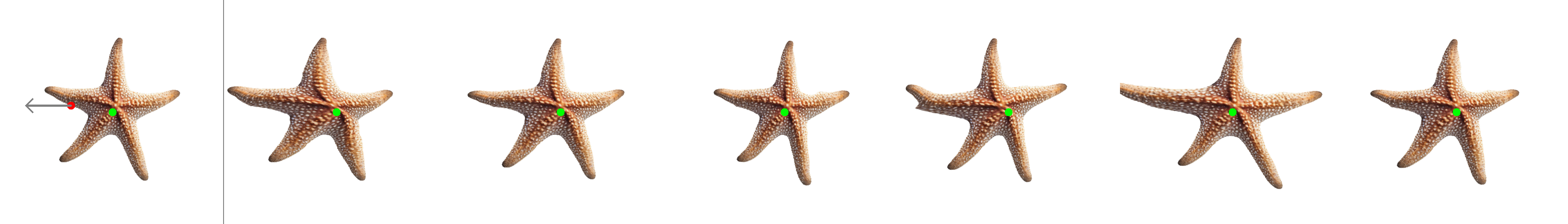} 
\Image{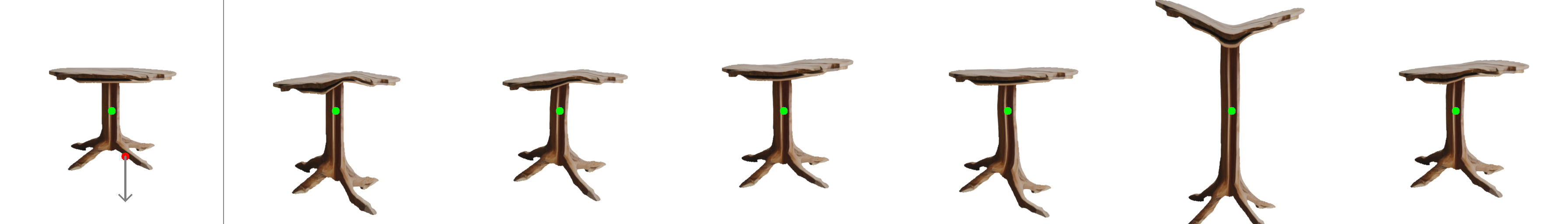}
}
\graphicspath{{Figures/supp/ablation/}}

\makeatletter
\def\Image#1{%
  \multicolumn{\LT@cols}{l}{\includegraphics[width=\textwidth]{#1}}\\
}
\makeatother

\renewcommand{\LTcaptype}{figure}
\LTcapwidth=\textwidth
\setlength{\tabcolsep}{0em}
\def\arraystretch{0.0}
\newcolumntype{Z}{>{\centering\arraybackslash}m{0.14\textwidth}}
{\scriptsize
\begin{longtable}[t!]
{ZZZZZZZ}
\captionsetup{type=figure, labelfont=bf}
\caption{\textbf{Ablation study for 2D mesh editing.} We examine the impact of each design choice on deformation outputs, including the use of diffusion prior (the second column), LoRA finetuning (the third column), two-stage pipeline (the fourth column), and initialization strategies during the \FirstStage{} (the fifth and sixth column).}\\
\label{fig:suppl_2d_ablation}\\
\makecell{Source} &
\makecell{$\mathcal{L}_h$ Only} &
\makecell{No LoRA~\cite{Hu:2022LoRA}} &
\makecell{\SecondStage{} Only} &
\makecell{ARAP Init.} &
\makecell{Poisson Init.} &
\makecell{Ours} \\
  \endhead

  \endfoot

  \Image{Images/2D/Supp_ablation/1.png} 
\Image{Images/2D/Supp_ablation/9.png} 
\Image{Images/2D/Supp_ablation/11.png} 
\Image{Images/2D/Supp_ablation/27.png} 
\Image{Images/2D/Supp_ablation/30.png} 
\Image{Images/2D/Supp_ablation/31.png} 
\Image{Images/2D/Supp_ablation/36.png} 
\Image{Images/2D/Supp_ablation/41.png} 
\Image{Images/2D/Supp_ablation/46.png} 
\Image{Images/2D/Supp_ablation/47.png} 
\Image{Images/2D/Supp_ablation/49.png} 
\Image{Images/2D/Supp_ablation/61.png} 
\Image{Images/2D/Supp_ablation/69.png} 
\Image{Images/2D/Supp_ablation/74.png}

\end{longtable}
}

\end{document}